\documentclass[lettersize,journal]{IEEEtran}
\usepackage{amsfonts}
\usepackage{amsmath}
\usepackage{algorithmic}
\usepackage{algorithm}
\usepackage{array}
\usepackage[caption=false,font=normalsize,labelfont=sf,textfont=sf]{subfig}
\usepackage{textcomp}
\usepackage{stfloats}
\usepackage{url}
\usepackage{verbatim}
\usepackage{graphicx}
\usepackage{cite}
\usepackage{tabularx} % for 'tabularx' env. and 'X' col. type
\usepackage{ragged2e} % for \RaggedRight macro
\usepackage{tabularx} % for 'tabularx' env. and 'X' col. type
\usepackage{booktabs} % for \toprule, \midrule etc macros
\usepackage{xspace}

\begin{document}

\title{FeaInfNet: Diagnosis in Medical Image with Feature-Driven Inference and Visual Explanations}

\author{
	\textbf{Yitao Peng\textsuperscript{1}},
	\textbf{Lianghua He\textsuperscript{1}},
	\textbf{Die Hu\textsuperscript{2*}},
	\textbf{Yihang Liu\textsuperscript{1}},
	\textbf{Longzhen Yang\textsuperscript{1}},
	\textbf{Shaohua Shang\textsuperscript{1}}\\
	\textsuperscript{1}School of Electronic and Information Engineering, Tongji University, Shanghai 201804, China\\
	\textsuperscript{2}School of Information Science and Technology, Fudan University, Shanghai 200433, China\\
	\{pyt, helianghua, 2111131, yanglongzhen, shaohuashang\}@tongji.edu.cn, hudie@fudan.edu.cn
	
}
%\author{Yitao Peng, Yihang Liu, Longzhen Yang, Lianghua He}
        % <-this % stops a space
%\thanks{This paper was produced by the IEEE Publication Technology Group. They are in Piscataway, NJ.}% <-this % stops a space
%\thanks{Manuscript received April 19, 2021; revised August 16, 2021.}}



\maketitle

\begin{abstract}
	Interpretable deep learning models have received widespread attention in the field of image recognition. Due to the unique multi-instance learning of medical images and the difficulty in identifying decision-making regions, many interpretability models that have been proposed still have problems of insufficient accuracy and interpretability in medical image disease diagnosis. To solve these problems, we propose feature-driven inference network (FeaInfNet). Our first key innovation involves proposing a feature-based network reasoning structure, which is applied to FeaInfNet. The network of this structure compares the similarity of each sub-region image patch with the disease templates and normal templates that may appear in the region, and finally combines the comparison of each sub-region to make the final diagnosis. It simulates the diagnosis process of doctors to make the model interpretable in the reasoning process, while avoiding the misleading caused by the participation of normal areas in reasoning. Secondly, we propose local feature masks (LFM) to extract feature vectors in order to provide global information for these vectors, thus enhancing the expressive ability of the FeaInfNet. Finally, we propose adaptive dynamic masks (Adaptive-DM) to interpret feature vectors and prototypes into human-understandable image patches to provide accurate visual interpretation. We conducted qualitative and quantitative experiments on multiple publicly available medical datasets, including RSNA, iChallenge-PM, Covid-19, ChinaCXRSet, and MontgomerySet. The results of our experiments validate that our method achieves state-of-the-art performance in terms of classification accuracy and interpretability compared to baseline methods in medical image diagnosis. Additional ablation studies verify the effectiveness of each of our proposed components.
\end{abstract}

\section{Introduction} \label{Section_1}
Deep learning technologies \cite{chandrasekaran2023retinopathy,song2023odspc} have made significant progress in recent years, with many algorithms showing higher accuracy than human experts in specific computer vision tasks \cite{bayoudh2021survey,chen2022boosting}. This trend has had a profound impact on the field of biomedical imaging. Deep learning algorithms have demonstrated excellent capabilities in classification, detection, and segmentation tasks in the field of biomedical imaging \cite{cheng2022contour,garcia2023secure}, providing doctors with support for manual diagnosis and decision-making. However, the application of these technologies in actual medical scenarios faces considerable challenges, one of the most prominent issues being trust \cite{gu2020vinet}. Since medical decisions may have a profound impact on patients' lives, medical diagnostic application models \cite{xi2020integrated,lin2023deep} must not only have excellent performance but also provide a strong basis for judgment. Therefore, designing a model that has both high recognition accuracy and good interpretability for medical diagnosis tasks has become an urgent problem that needs to be solved.

\begin{figure}[!t]
	\centerline{\includegraphics[width=\columnwidth]{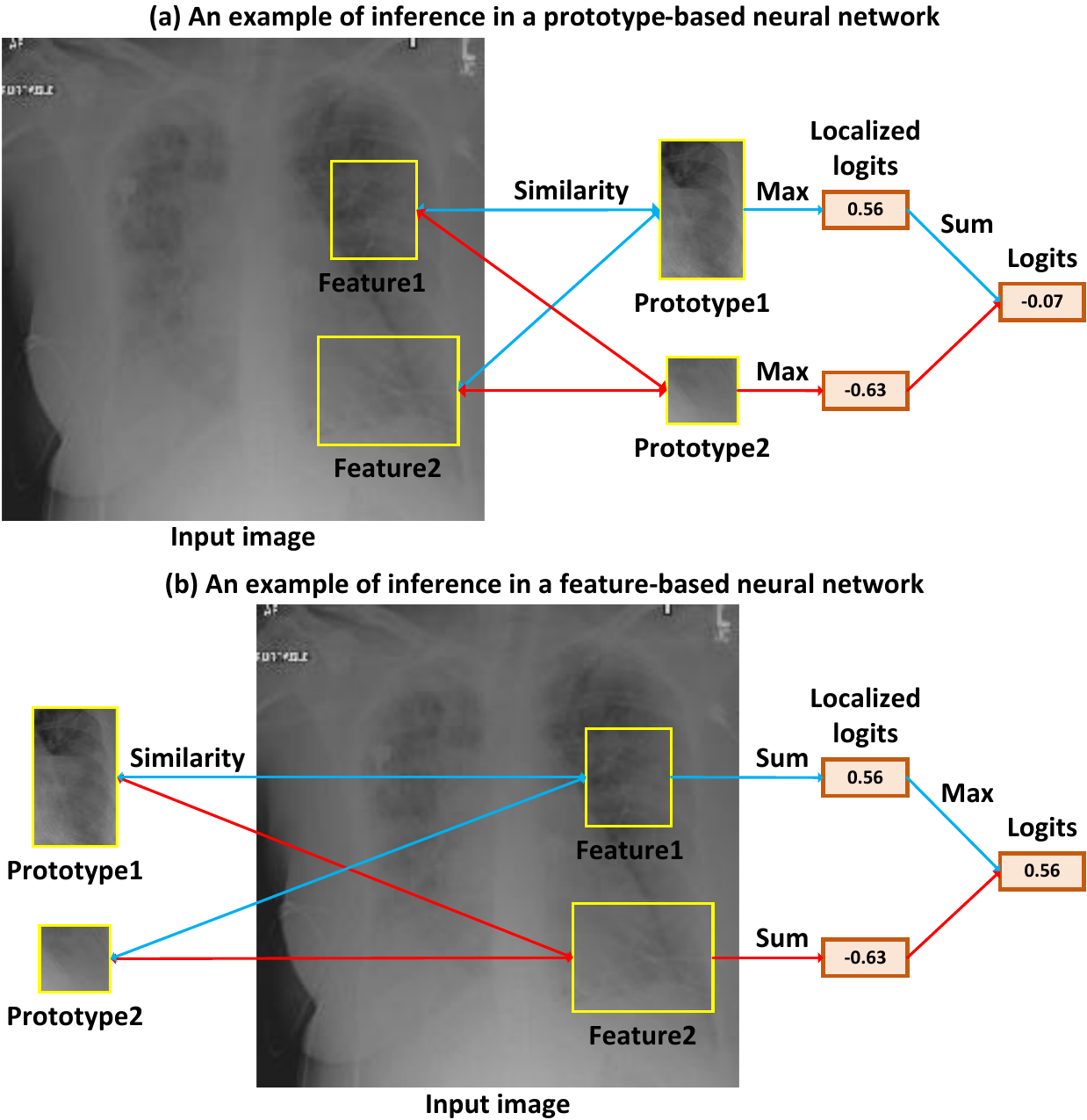}}
	\caption{Schematic diagram of the inference process of prototype-based neural network and feature-based neural network for diagnosing medical images.}
	\label{fig_1}
\end{figure}

There have been many studies trying to improve the interpretability of deep learning models in medical image diagnosis tasks. Perturbation-based methods \cite{fong2017interpretable,yuan2020interpreting} analyze how perturbations in the input affect the model's predictions. Gradient-based methods \cite{shrikumar2017learning,sundararajan2017axiomatic} generate saliency maps by computing the derivatives of model class scores with respect to the input image. Methods based on class activation mapping \cite{selvaraju2017grad,jiang2021layercam} utilize gradient information to generate category-specific saliency maps from input images to provide visual explanations for the model's prediction results. Although these traditional methods are used in many areas of medical imaging. However, these methods are post-hoc analyses. They only provide the area where the model focuses in the medical image and cannot explain the way the model predicts \cite{rudin2019stop}. Therefore, people still cannot fully trust the model's decision-making.

In order to make the model itself interpretable, prototype-based neural networks \cite{rymarczyk2022interpretable,donnelly2022deformable} have become a promising approach recently. It encodes the input image into feature maps, and then extracts a local feature vector from the feature maps to compare the similarity with the template of each category (called a prototype) to make a classification. This reasoning structure is used to simulate human behavior. The analysis process realizes the interpretability of the model reasoning process. At the same time, it explains the feature vectors and prototypes used for decision-making by upsampling similarity activations, so that the network obtains visual level interpretability. Prototype-based neural networks \cite{chen2019looks,keswani2022proto2proto} are equipped with specific (1) inference structures, (2) feature extraction methods, and (3) saliency map generation methods for visual interpretation. This enables them to achieve outstanding performance in interpretable classification on natural images. However, there are still shortcomings in terms of recognition accuracy and interpretability in the context of interpretable diagnosis in medical imaging.

In the inference structure, prototype-based neural networks all adopt the reasoning structure shown in Figure \ref{fig_1} (a). They use each prototype as a benchmark, find the most similar features to the prototype in the input image to calculate the similarity score of the prototype to the input image, and then combine the similarity scores generated by all prototypes for classification. This interpretable reasoning structure is not suitable for medical imaging diagnosis. Because a medical image to be diagnosed may contain features of two categories at the same time, that is, there are both disease areas and normal areas in a disease image. Therefore, we do not conclude that the image is normal by finding that the normal features are similar to the normal prototype or that the normal features are dissimilar to the disease prototype. The prototype-based reasoning structure will integrate the similarity scores of normal areas to reduce the impact of the similarity scores of disease areas on the predicted logits, ultimately misleading the judgment of disease.

In terms of feature extraction method, the first proposed ProtoPNet \cite{chen2019looks} uses the $1 \times 1$ patch in the feature maps as the feature vector. Gen-ProtoPNet \cite{singh2021interpretable} generalizes the structure of the feature vector to a structure of any integer size in the feature maps. Due to the inductive bias of convolutional neural networks (CNNs), the feature vectors extracted by these networks represent local information in the image. This results in feature vectors and prototypes being unable to learn global information, thus limiting their expressive capabilities, which in turn leads to limited classification performance of the network.

For the saliency map generation method, since the prototype-based neural network adopts CNN as the backbone \cite{nauta2021neural,rymarczyk2021protopshare}, the similarity activations generated by comparing feature vectors with prototypes contain spatial positional information of the original image. Therefore, past methods generate saliency maps by upsampling similarity activations to interpret the regions represented by feature vectors and prototypes. However, similarity activations are relatively rough, and in medical images, the lesion areas used for decision-making are often very subtle, and the saliency map generated by upsampling similarity activations cannot accurately locate the pathological area used for decision-making.

In this paper, in order to solve the above three problems, we propose FeaInfNet, a model with high recognition performance and good interpretability dedicated to medical image diagnosis. The model works as follows:

(1) We propose a feature-based reasoning structure as shown in Figure \ref{fig_1} (b) to solve the misleading problem of prototype-based reasoning in medical imaging. FeaInfNet preserves the interpretable similarity comparison structure. It uses each feature of the image as a benchmark, compares each feature with the disease prototype and normal prototype that may appear at the location of the feature to generate a similarity score, and uses the largest similarity score among all features to classify. This method of relying only on a single sub-region feature for decision-making avoids the misleading problem caused by the simultaneous participation of disease features and normal features in different regions in the medical image in inferential diagnosis.

(2) We propose local feature masks (LFM) to optimize the traditional rigid feature vector extraction method. LFM extracts local information of the feature maps and supplements global information, which enhances the expressive ability of feature vectors and prototypes, thereby improving the classification performance of the network.

(3) Based on a powerful saliency map generation method dynamic masks learning (DM) \cite{10314012}, we propose  adaptive dynamic masks (Adaptive-DM) to replace the traditional upsampling similarity activations method to generate saliency maps for FeaInfNet and prototype-based neural networks to provide visual explanations. Based on DM, we propose adaptive weight learning to autonomously learn the weights between similarity terms and mask terms in DM. The consistent activation loss of DM can well weigh the importance between similarity items and mask items. Let the saliency map most accurately retain the areas of most concern for network decision-making, while removing redundant areas that are irrelevant for decision-making. This enables FeaInfNet and prototype-based neural networks to have better visual interpretability.

Our key contributions are as follows:

\begin{itemize}
	\item We propose a feature-based reasoning structure that retains the interpretability of the reasoning process while avoiding misunderstandings caused by normal areas in medical images participating in reasoning, thereby improving the accuracy of the network.
	\item We proposed LFM to extract feature vectors, so that the feature vectors retain local information while supplementing global information, enhance the expressive ability of feature vectors and prototypes, and thereby improve the classification accuracy of the network.
	\item We proposed an adaptive weight learning based on DM to form Adaptive-DM, and replaced the traditional upsampling similarity activations to provide better visual explanations for FeaInfNet and prototype-based networks.
\end{itemize}

\begin{figure*}[!t]
	\centering
	{\includegraphics[width=1.0\linewidth]{{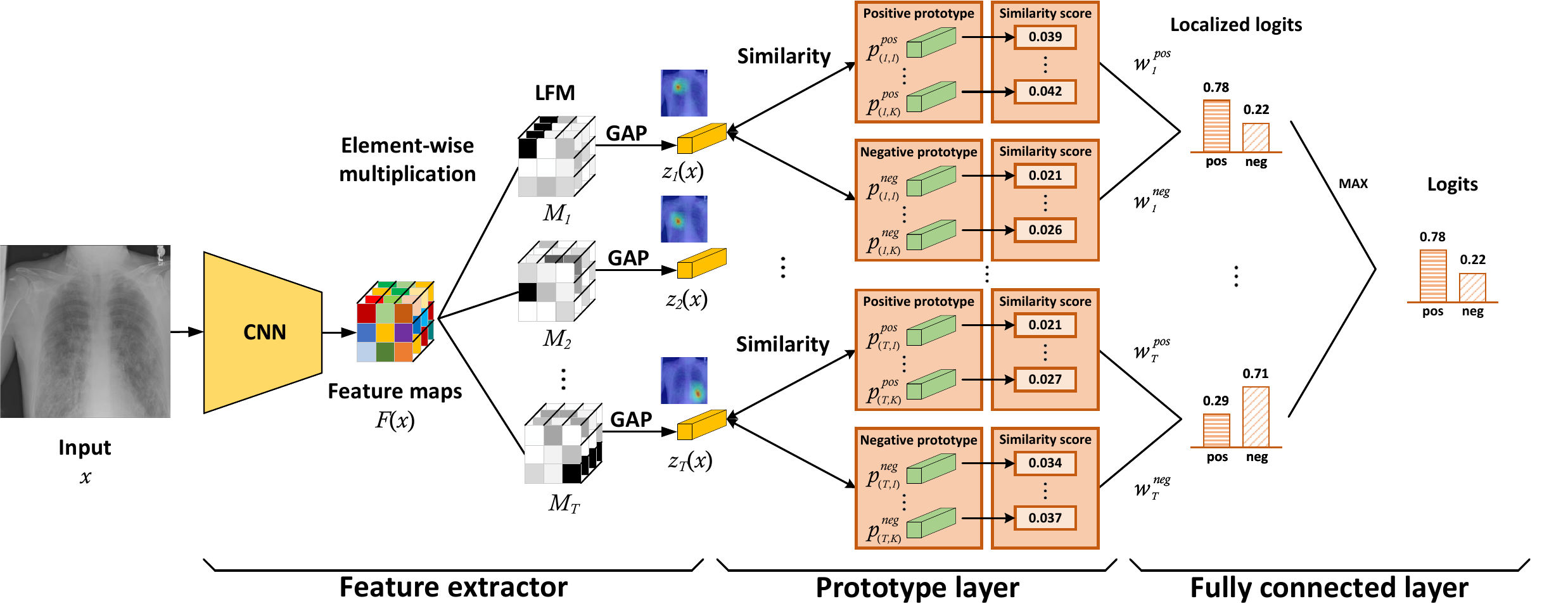}}%
	}
	\caption{Overview of the FeaInfNet architecture. First, the feature extractor encodes the image, generates feature maps and extracts feature vectors. Subsequently, the prototype layer compares these feature vectors with the corresponding prototypes. Finally, the fully connected layer utilizes the compared similarity scores to classify the images.}
	\label{fig_overall}
\end{figure*}

\section{Related Work} \label{Section_2}
\subsection{Attribution Methods} \label{Section_2_1}
Attribution methods generate visually interpreted saliency maps to represent the importance of each pixel of an image to the model's classification. It is one of the most popular techniques for explaining model decisions and is widely used in interpretability research in medical diagnosis. Class activation mapping (CAM) \cite{zhou2016learning} uses a linear combination of the outputs of the last global average pooling layer of the network to generate saliency maps. CheXNeXt \cite{rajpurkar2018deep} was proposed to detect lung pathology and use CAM to identify locations on chest radiographs that contribute most to the model predictions. Grad-CAM \cite{selvaraju2017grad} uses the gradients flowing into the final convolutional layer to generate localization maps that highlight important regions in the image for predictive concepts, and it is used to study disease regions that the model considers more discriminative \cite{lin2020covid}. DeepLIFT \cite{shrikumar2017learning} was used to explain the decisions of multiple sclerosis classification models \cite{lopatina2020investigation}. IG \cite{sundararajan2017axiomatic} provides visual explanation for the task of predicting diabetic retinopathy from retinal fundus images \cite{sayres2019using}.

\subsection{Interpretable Models} \label{Section_2_2}
Learning prototypes during the model training phase makes the model itself interpretable for inference.The prototype-based network ProtoPNet \cite{chen2019looks} encodes the image into a feature map, in which the $1 \times 1$ patch feature vector is extracted and compared with the prototype for classification. Based on this again, Gen-ProtoPNet \cite{singh2021interpretable} generalizes the $1 \times 1$ size of the feature vector to a feature vector whose length and width are any integer size, thereby improving the expression ability of the feature vector. XProtoNet \cite{kim2021xprotonet} uses occurrence maps to learn the unique position of each disease in X-rays to provide local and global explanations for medical diagnosis. NP-ProtoPNet \cite{singh2021these} introduces negative reasoning by fixing the weight of the classification layer to improve the network's recognition ability in medical images. TesNet \cite{wang2021interpretable} bridges high-level input patches and output categories by introducing a plug-in transparent embedding space to design interpretable neural networks.

Previous prototype-based networks \cite{chen2019looks,rymarczyk2022interpretable} achieved good accuracy and interpretability in natural image recognition tasks, but their performance was still insufficient for medical image diagnosis. Therefore, we proposed FeaInfNet to provide a model with high classification accuracy and strong interpretability for medical image diagnosis.

\section{Methodology} \label{Section_3}

This section provides an overview of the feature-based inference and training process of FeaInfNet (Section \ref{Inference and Training of FeaInfNet}), describes the method for extracting feature vectors (Section \ref{Extraction LFM}), and introduces the working principles of Adaptive-DM for visualizing feature vectors and prototypes (Section \ref{Adaptive-DM}).

\subsection{Inference and Training of FeaInfNet} \label{Inference and Training of FeaInfNet}

\textbf{Recognition Process.} The overall structure of FeaInfNet is shown in Figure \ref{fig_overall}. It makes the network interpretable in the reasoning process by simulating the doctor's disease diagnosis process. Its calculation process is briefly described as follows: it first encodes the input image into the feature maps through CNN, and extracts feature vectors through LFM to represent the information about the positions of different areas of the input image. Secondly, it compares the similarity between the feature vector and the learned disease prototype and normal prototype template sets at the corresponding positions to obtain the similarity score. Then, the similarity scores are combined to calculate local logits. Finally, the logits with the highest probability of disease are selected as predictions. Next we describe the above calculation process in detail.

Let the input image be $x \in R^{H \times W \times C}$, where $H$, $W$, and $C$ are the height, width, and number of channels of the input image respectively. $x$ is encoded by CNN to obtain the feature maps $F(x) \in R^{H_{1} \times W_{1} \times C_{1}}$, where $H_{1}$, $W_{1}$, and $C_{1}$ are the height, width, and number of channels of the feature maps respectively. FeaInfNet extracts feature vectors $\{z_{t}(x)\}_{t=1}^{T}$ in $T$ ($T = H_{1} \times W_{1}$) different areas (the specific extraction method is introduced in Section \ref{Extraction LFM}). These $T$ feature vectors $z_{t}(x)$ represent the image patch information at different positions in the input image $x$.

The prototype represents the image patch with specific category features learned by FeaInfNet from the training set. For medical images used to diagnose a certain disease, normal areas in different locations exhibit different characteristics. For example, the characteristics of the heart and stomach are different in normal people, but disease areas will have the same disease characteristics even if they are in different locations in the image. Therefore, we make the disease prototypes learned by FeaInfNet shared, but the normal prototypes not shared. Specifically, we set shared disease prototypes $\{ p^{pos}_{j} \}^{K^{pos}}_{j=1}$ to represent possible disease characteristics in this type of image, and normal prototypes $\{ p^{neg}_{(t, j)} \}^{K^{neg}_{t}}_{j=1}$ to represent the normal organ characteristics of normal people in $t$ regions. $p^{pos}_{j}$ represents the disease features in this type of image, and $p^{neg}_{(t,j)}$ represents the normal features of this type of image in area $t$.

We define the similarity scores between the disease prototype $p^{pos}_{j}$ and the normal prototype $p^{neg}_{(t,j)}$ with the feature vector $z_{t}(x)$ as follows:
\begin{equation}
\begin{aligned}
g^{pos}_{(t,j)}(x) = log(\frac{||z_{t}(x)-p^{pos}_{j}||^{2}+1}{||z_{t}(x)-p^{pos}_{j}||^{2}+\epsilon})
\end{aligned}
\end{equation}
\begin{equation}
\begin{aligned}
g^{neg}_{(t,j)}(x) = log(\frac{||z_{t}(x)-p^{neg}_{(t,j)}||^{2}+1}{||z_{t}(x)-p^{neg}_{(t,j)}||^{2}+\epsilon})
\end{aligned}
\end{equation}
where $\epsilon$ is a small positive constant to avoid division by zero. $g^{pos}_{(t,j)}(x)$ and $g^{neg}_{(t,j)}(x)$ respectively represent the degree of similarity between the image patch represented by the feature vector $z_{t}(x)$ and the image patch represented by the disease prototype $p^{pos}_{j}$ and the normal prototype $p^{neg}_{(t,j)}$. We note that $y=0$ and $y=1$ indicate that the input image is predicted to be normal and predicted to be diseased, respectively. The disease prediction probability $P_{t}(y=1|x)$ in the region $t$ of the input image $x$ is as follows:
\begin{equation}
\begin{aligned}
P_{t}(y=1|x) = \sum_{j=1}^{K^{pos}}|w^{pos}_{(t,j)}|g^{pos}_{(t,j)}(x) - \sum_{j=1}^{K^{neg}}|w^{neg}_{(t,j)}|g^{neg}_{(t,j)}(x)
\end{aligned}
\end{equation}

Based on the predicted probabilities of $t$ regions, the predicted probability in the region $t$ with the highest disease probability is selected and normalized to the final disease probability $P(y=1|x)$. The formula is as follows:
\begin{equation}
\begin{aligned}
P(y=1|x) = \frac{e^{\mathop{\max}\limits_{1 \leq t \leq T}P_{t}(y=1|x)}}{e^{\mathop{\max}\limits_{1 \leq t \leq T}P_{t}(y=1|x)} + e^{-\mathop{\max}\limits_{1 \leq t \leq T}P_{t}(y=1|x)}}
\end{aligned}
\end{equation}

The probability of normal is $P(y=0|x)$. 
\begin{equation}
\begin{aligned}
P(y=0|x) = 1 - P(y=1|x)
\end{aligned}
\end{equation}

If $P(y=1|x) > P(y=0|x)$, we judge it to be a disease. Compared with the prototype-based reasoning structure, this feature-based reasoning structure can avoid the misleading caused by normal areas, thereby improving the accuracy of network. We provide the mathematical proof in Appendix.

\textbf{Training Scheme.} In this section we introduce how to use medical imaging data sets to train FeaInfNet. Define the training data set as $\{(x_{h}, y_{h})\}^{n_{h}}_{h=1}$, where $x_{h}$ is the training image and $y_{h}$ is the corresponding label. The disease train datasets $\{(x^{pos}_{h}, y^{pos}_{h})\}^{n^{pos}_{h}}_{h=1}$, where $y^{pos}_{h} = 1$. The normal train datasets $\{(x^{neg}_{h}, y^{neg}_{h})\}^{n^{neg}_{h}}_{h=1}$, where $y^{neg}_{h} = 0$. $n_{h} = n^{pos}_{h} + n^{neg}_{h}$. FeaInfNet sets $T$ regions, with disease prototypes $\{ p^{pos}_{j} \}^{K^{pos}}_{j=1}$ and normal prototypes $\{ p^{neg}_{(t, j)} \}^{K^{neg}_{t}}_{j=1}$.

In order to deal with the imbalance in the number of disease and normal samples in the medical imaging data set, we adopt a weighted balance loss to train FeaInfNet, as follows:
\begin{equation}
\begin{aligned}
H(P(x_{h}), y_{h}) = & - (1-P(x_{h}))^{\kappa}y_{h}log(P(x_{h})) \\
& - (P(x_{h}))^{\kappa}(1 - y_{h})log(1-P(x_{h}))
\end{aligned}
\end{equation}
where $P(x_{h}) = P(y=1|x_{h})$ and $y_{h} \in \{0, 1\}$. To better learn about disease and normal prototypes. We define the clustering cost minimization losses $L^{neg}_{Clst}$ and $L^{pos}_{Clst}$, and the separation cost minimization losses $L^{neg}_{Sep}$ and $L^{pos}_{Sep}$.

When a disease image $x^{pos}_{h}$ is input, the feature vector $z_{t}(x^{pos}_{h})$ corresponding to at least one area in the image $x^{pos}_{h}$ is close to at least one disease prototype $p^{pos}_{j}$.
\begin{equation}
\begin{aligned}
L^{pos}_{Clst} = \frac{1}{n^{pos}_{h}} \sum_{h=1}^{n^{pos}_{h}}\mathop{\min}\limits_{1 \leq t \leq T}\mathop{\min}\limits_{1 \leq j \leq K^{pos}_{t}}||z_{t}(x^{pos}_{h})-p^{pos}_{j}||^{2}
\end{aligned}
\end{equation}

When a normal image $x^{neg}_{h}$ is input, the feature vectors $\{z_{t}(x^{neg}_{h})\}_{t=1}^{T}$ of all regions are at least close to a normal prototype $p^{neg}_{(t,j)}$ under their corresponding regions.
\begin{equation}
\begin{aligned}
L^{neg}_{Clst} = \frac{1}{n^{neg}_{h}} \sum_{h=1}^{n^{neg}_{h}}\mathop{\max}\limits_{1 \leq t \leq T}\mathop{\min}\limits_{1 \leq j \leq K^{neg}_{t}}||z_{t}(x^{neg}_{h})-p^{neg}_{(t,j)}||^{2}
\end{aligned}
\end{equation}

When a normal images $x^{neg}_{h}$ is input, the feature vectors $\{z_{t}(x^{neg}_{h})\}_{t=1}^{T}$ corresponding to all regions are far away from all disease prototypes $\{ p^{pos}_{j} \}^{K^{pos}}_{j=1}$.
\begin{equation}
\begin{aligned}
L^{neg}_{Sep} = - \frac{1}{n^{neg}_{h}} \sum_{h=1}^{n^{neg}_{h}}\mathop{\min}\limits_{1 \leq t \leq T}\mathop{\min}\limits_{1 \leq j \leq K^{pos}}||z_{t}(x^{neg}_{h})-p^{pos}_{j}||^{2}
\end{aligned}
\end{equation}

When a disease image $x^{pos}_{h}$ is input, the feature vector $z_{t}(x^{pos}_{h})$ of at least one region is far away from the normal prototype $p^{neg}_{(t,j)}$ of the region corresponding to $z_{t}(x^{pos}_{h})$.
\begin{equation}
\begin{aligned}
L^{pos}_{Sep} = - \frac{1}{n^{pos}_{h}} \sum_{h=1}^{n^{pos}_{h}}\mathop{\max}\limits_{1 \leq t \leq T}\mathop{\max}\limits_{1 \leq j \leq K^{neg}_{t}}||z_{t}(x^{pos}_{h})-p^{neg}_{(t,j)}||^{2}
\end{aligned}
\end{equation}

The total loss function is defined as follows:
\begin{equation}
\begin{aligned}
L = & \frac{1}{n_{h}} \sum_{h=1}^{n_{h}} H(P(x_{h}), y_{h}) + \eta_{1}L^{neg}_{Clst} \\ 
& + \eta_{2}L^{neg}_{Sep} + \eta_{3}L^{pos}_{Clst} + \eta_{4}L^{pos}_{Sep}
\end{aligned}
\end{equation}
where $\{\eta_{i}\}_{i=1}^{4}$ are hyperparameters.

\textbf{Prototype Learning.}
In order to make each disease prototype and normal prototype correspond to a specific medical image patch, so that people can more intuitively understand the information represented by the prototype. We traverse the entire training data set and project $p^{pos}_{j}$ and $p^{neg}_{(t,j)}$ onto the image patches most similar to them in all disease images $\{x^{pos}_{h}\}^{n^{pos}_{h}}_{h=1}$ and all normal images $\{x^{neg}_{h}\}^{n^{neg}_{h}}_{h=1}$ respectively.
\begin{equation}
\begin{aligned}
p^{pos}_{j} \gets \mathop{\arg\min}\limits_{z_{t}(x^{pos}_{h})}||z_{t}(x^{pos}_{h}) - p^{pos}_{j} ||
\end{aligned}
\end{equation}
\begin{equation}
\begin{aligned}
p^{neg}_{(t,j)} \gets \mathop{\arg\min}\limits_{z_{t}(x^{neg}_{h})}||z_{t}(x^{neg}_{h}) - p^{neg}_{(t,j)} ||
\end{aligned}
\end{equation}
where $t \in \{1, 2, ..., T\}$. $z_{t}(x^{pos}_{h})$ and $z_{t}(x^{neg}_{h})$ respectively represent the feature vectors extracted from the image $x^{pos}_{h}$ and $x^{neg}_{h}$ (their definitions are shown in Equation (\ref{LFM_extract})). Through the above prototype learning, each prototype can be understood as a specific medical image patch, so that they can be interpreted more intuitively.

\subsection{Extraction of Feature with the Local Feature Masks} \label{Extraction LFM}

\begin{figure}[!t]
	\centerline{\includegraphics[width=\columnwidth]{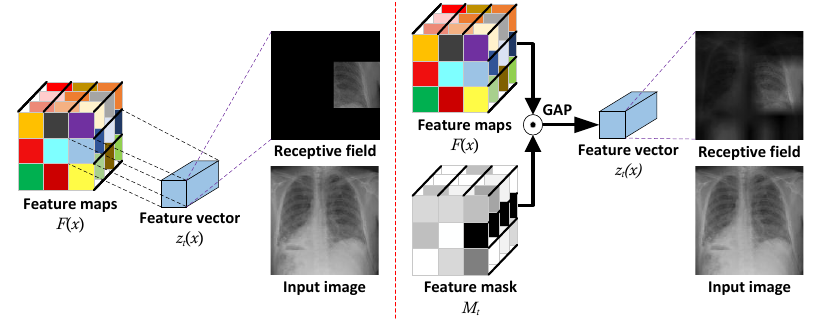}}
	\caption{The left and right sides show the calculation process of feature extraction the prototype-based neural network and FeaInfNet respectively.}
	\label{prototype_vector}
\end{figure}

Traditional prototype-based networks \cite{chen2019looks,wang2021interpretable} utilize local convolutions of CNN to encode input images into feature maps. The elements of all channels at each position in the feature maps represent information about the local area of the corresponding receptive field in the image. As shown on the left side of Figure \ref{prototype_vector}, the previous prototype-based network extracts elements of the $1 \times 1 \times C_{1}$ rigid structure from the feature maps as feature vectors to participate in subsequent reasoning. This feature vector only learns the local area covered by adjacent convolution kernels and ignores global information, which is especially unfavorable to the network's learning of medical images.

Several works have shown that global information can provide beneficial guidance and constraints for the local feature extraction process \cite{wang2020non,yang2022focal}. Therefore, we propose local feature masks (LFM) to extract feature vectors, as shown on the right side of Figure \ref{prototype_vector}. This enables the feature vector to learn local and global information at the same time, and the network can more comprehensively understand the local and overall correlation of the input data to improve the expression ability of the feature vector and the classification accuracy of the network.

Specifically, we assume that the feature maps extracted by CNN is $F(x) \in R^{H_{1} \times W_{1} \times C_{1}}$. Set LFM to consist of $T=H_{1} \times W_{1} $ feature masks $\{M_{t}\}_{t=1}^{H_{1} \times W_{1}}$, where $M_{t} \in R^{H_{1} \times W_{1} \times C_{1}}$ is expressed in matrix form as follows:
\begin{equation}
\begin{aligned}
M_{t} = [m^{t}_{i, j}]_{H_{1} \times W_{1}}
\end{aligned}
\end{equation}
\begin{equation}
\begin{aligned}
m^{t}_{i, j} = \begin{cases} E + \alpha^{t}_{i,j}[U(0,1), ..., U(0,1)] , (i, j) = (\phi_{t}, \varphi_{t}) \\  \alpha^{t}_{i,j}[U(0,1), ..., U(0,1)], \  (i, j) \neq (\phi_{t}, \varphi_{t}) 
\end{cases}
\end{aligned}
\end{equation}
where $\phi_{t} = \left \lfloor \frac{t}{W_{1}} \right \rfloor, \varphi_{t} = t - W_{1}\left \lfloor \frac{t}{W_{1}} \right \rfloor $, $\alpha^{t}_{i,j}$ is the hyperparameters, $U(0,1)$ represents a $0-1$ uniform distribution, and $E$ is a vector of length $C_{1}$ and each element is 1. The feature masks $\{M_{t}\}_{t=1}^{H_{1} \times W_{1}}$ of LFM focus on pairwise distinct subregions of $F(x)$, covering every subregion of the feature maps. As shown on the right side of Figure \ref{prototype_vector}, FeaInfNet extracts feature vectors by multiplying LFM $M_{t}$ and feature maps $F(x)$ element by element and then taking global average pooling (GAP), instead of the method of extracting rigid feature vectors shown on the left side of Figure \ref{prototype_vector}. The mathematical formula is as follows:
\begin{equation} \label{LFM_extract}
\begin{aligned}
z_{t}(x) = GAP(M_{t}F(x))
\end{aligned}
\end{equation}

\subsection{Adaptive Dynamic Masks} \label{Adaptive-DM}
Previous prototype-based networks generate similarity activations by comparing the prototype with the feature vector at each location on the feature map. Because each position in the similarity activation represents the degree of similarity between the image patch corresponding to the feature vector and the image patch corresponding to the prototype. Therefore, they generate saliency maps by upsampling similarity activations to the input image size to display the regions represented by feature vectors and prototypes, providing visual explanations for the network's decisions. Due to the rough division of similarity activation, the decision-making areas emphasized by the saliency map generated by upsampling are not refined, especially in the analysis of subtle lesions in medical images, which are inaccurately interpreted.

DM is the most advanced neural network visual interpretation method recently proposed. It uses mask vector upsampling of different sizes to perturb the input image and observe the output of the detection node, and trains a fine saliency map by optimizing the consistent activation loss composed of mask terms and similarity terms. However, the weight parameters between the mask term and the similarity term in the previous consistent activation loss of DM were manually defined, which could not optimally weigh the mask term and the similarity term, resulting in inaccurate saliency maps generated. To solve this problem, we propose Adaptive-DM, which uses an adaptive weight learning method to analyze the most appropriate weight parameters to generate high-quality saliency maps. Finally, we set the feature vector and prototype as the detection node of Adaptive-DM. This replaces the upsampling similarity activations method for FeaInfNet or other prototype-based neural networks to interpret the image patches represented by feature vectors and prototypes, thereby providing accurate visual explanations. The Adaptive-DM consists of two operations: dynamic masks learning and adaptive weight learning. The following provides an introduction to these components:

\subsubsection{Dynamic Masks Learning}
The process of DM is briefly described as three steps: (1) prepare the mask vector to be learned, (2) determine the detection node of the network, and (3) train the mask vector by constraining the detection node to have consistent activation. The learning are as follows.

(1) Mask vectors $\{\delta_{i}\}^{N_{d}}_{i=1}$ to be learned, where $\delta_{i} \in R^{u_{i} \times v_{i}}$, represent mask vectors of different sizes, with $u_{i}$ and $v_{i}$ denoting the height and width of the mask vector. For any $i,j\in\{1,2,...,N_{d}\}$, if $i \neq j$, then $u_{i} \neq u_{j}$ or $v_{i} \neq v_{j}$. $\delta_{i}$ is initialized as $\delta^{0}_{i}$, where each element of $\delta^{0}_{i}$ is initialized to a fixed value $\xi$. For a mask vector of size $u_{i} \times v_{i}$, it divides the input image into $u_{i} \times v_{i}$ subregions, each with a size of $\frac{H}{u_{i}} \times \frac{W}{v_{i}}$, where $H$ and $W$ are the height and width of the input image. By learning from mask vectors of different sizes, the network analyzes the importance of each subregion under various size divisions. The upsampling function $g(\cdot)$, where $g(\delta_{i}) \in R^{H \times W \times 1}$, is used to mask the image to train $\delta_{i}$.

(2) It is set that the feature vectors and prototypes in FeaInfNet are used as detection nodes to interpret the image patches represented by the feature vectors and prototypes. Therefore, the location of the feature vector and prototype is found through the following indexes. Let the input image be $x^{0}$, the interpreted feature vector and prototype are $z_{t}(x^{0})$ and $p_{j}$ respectively, and the image that generates $p_{j}$ after prototype learning is $x^{p_{j}}$. Below, we use Adaptive-DM to analyze the decision-making area (image patch) corresponding to the feature vector $z_{t}(x^{0})$ and the prototype $p_{j}$ in $x^{0}$ and $x^{p_{j}}$. Note that the feature mask indexes that generate $z_{t}(x^{0})$ and $p_{j}$ are $t_{x^{0}}$ and $t_{p_{j}}$ respectively. Based on these indexes, the locations of detection nodes (i.e. $z_{t}(x^{0})$ and $p_{j}$) in the network are found based on these indices.
\begin{equation}
t_{x^{0}}  = \ \mathop{argmax}\limits_{t}\ P_{t}(y=1|x^{0})
\end{equation}
\begin{equation}
t_{p_{j}}  = \ \mathop{argmin}\limits_{t}\ ||z_{t}(x^{p_{j}})-p_{j}||_{2}
\end{equation}
(3) Let $\{\delta^{x^{0}}_{i}\}^{N_{d}}_{i=1}$ and $\{\delta^{x^{p_{j}}}_{i}\}^{N_{d}}_{i=1}$ are the mask vectors of the feature vector $z_{t}(x^{0})$ and the prototype $p_{j}$ respectively. We define the similarity terms $Sim(\delta^{x^{0}}_{i})$ and $Sim(\delta^{x^{p_{j}}}_{i})$ of $\delta^{x^{0}}_{i}$ and $\delta^{x^{p_{j}}}_{i}$ respectively as the square of the difference between the output values of the detection node when the original image and the mask image are input into the network at the same time:
\begin{equation}
\begin{aligned}
Sim(\delta^{x^{0}}_{i}) = ||z_{t_{x^{0}}}(x^{0})-z_{t_{x^{0}}}(g(\delta^{x^{0}}_{i})x^{0})||^{2}_{2}
\end{aligned}
\end{equation}
\begin{equation}
\begin{aligned}
Sim(\delta^{x^{p_{j}}}_{i}) = ||z_{t_{p_{j}}}(x^{p_{j}})-z_{t_{p_{j}}}(g(\delta^{x^{p_{j}}}_{i})x^{p_{j}})||^{2}_{2}
\end{aligned}
\end{equation}

Define mask terms $Mas(\delta^{x^{0}}_{i})$ and $Mas(\delta^{x^{p_{j}}}_{i})$ as the L1 regularization values of mask vectors $\delta^{x^{0}}_{i}$ and $\delta^{x^{p_{j}}}_{i}$ respectively divided by their own height and width.
\begin{equation}
\begin{aligned}
Mas(\delta^{x^{0}}_{i}) = \frac{||\delta^{x^{0}}_{i}||_{1}}{|u_{i}^{x^{0}}v_{i}^{x^{0}}|}
\end{aligned}
\end{equation}
\begin{equation}
\begin{aligned}
Mas(\delta^{x^{p_{j}}}_{i}) = \frac{||\delta^{x^{p_{j}}}_{i}||_{1}}{|u_{i}^{x^{p_{j}}}v_{i}^{x^{p_{j}}}|}
\end{aligned}
\end{equation}

Define the consistent activation losses $Con(\delta^{x^{0}}_{i})$ and $Con(\delta^{x^{p_{j}}}_{i})$ of $\delta^{x^{0}}_{i}$ and $\delta^{x^{p_{j}}}_{i}$ as the weighted sum of their own similarity terms and mask terms.
\begin{equation}
\begin{aligned}
Con(\delta^{x^{0}}_{i}) = Sim(\delta^{x^{0}}_{i}) + \lambda_{i} Mas(\delta^{x^{0}}_{i})
\end{aligned}
\end{equation}
\begin{equation}
\begin{aligned}
Con(\delta^{x^{p_{j}}}_{i}) = Sim(\delta^{x^{p_{j}}}_{i}) + \lambda_{i} Mas(\delta^{x^{p_{j}}}_{i})
\end{aligned}
\end{equation}

The mask vectors $\{\delta^{x^{0}}_{i}\}^{N_{d}}_{i=1}$ and $\{\delta^{x^{p_{j}}}_{i}\}^{N_{d}}_{i=1}$ are trained by minimizing $Con(\delta^{x^{0}}_{i})$ and $Con(\delta^{x^{p_{j}}}_{i})$, so that $\{\delta^{x^{0}}_{i}\}^{N_{d}}_{i=1}$ and $\{\delta^{x^{p_{j}}}_{i}\}^{N_{d}}_{i=1}$ retain decision-related areas and eliminate decision-irrelevant areas. Finally, the trained mask vectors $\{\delta^{x^{0}}_{i}\}^{N_{d}}_{i=1}$ and $\{\delta^{x^{p_{j}}}_{i}\}^{N_{d}}_{i=1}$ are upsampled and stacked, and the noise areas are removed to obtain the saliency maps $S^{x^{0}}$ and $S^{x^{p_{j}}}$ represented by the feature vector $z_{t}(x^{0})$ and prototype $p_{j}$. The formula are as follows:
\begin{equation} \label{Ax}
S^{x^{0}} = N(\{\sum^{N_{d}}_{i=1}g(\delta^{x^{0}}_{i}) \geq \Omega \}(\sum^{N_{d}}_{i=1}g(\delta^{x^{0}}_{i}) - \Omega))
\end{equation}
\begin{equation} \label{Axb}
S^{x^{p_{j}}} = N(\{\sum^{N_{d}}_{i=1}g(\delta^{x^{p_{j}}}_{i}) \geq \Omega \}(\sum^{N_{d}}_{i=1}g(\delta^{x^{p_{j}}}_{i}) - \Omega))
\end{equation}
where the threshold matrix $\Omega = [\omega]_{H \times W}$ and threshold value $\omega$ are used to remove noise information
in the saliency maps. $\{\cdot \}$ represents a truth function that equals 1 when the statement is true and 0 when it is false. $N(B)$ is a normalization function that normalizes ($\frac{\beta_{i,j} - min(B)}{max(B)-min(B)}$) each value $\beta_{i,j}$ in $B$, where $ B =[\beta_{i,j}]_{H \times W}$ is a matrix, and $max(B)$ and $min(B)$ are the maximum and minimum of $B$, respectively.

\subsubsection{Adaptive Weight Learning} \label{Adaptive_Weight_Learning}
\begin{figure}[!t]
	\centerline{\includegraphics[width=\columnwidth]{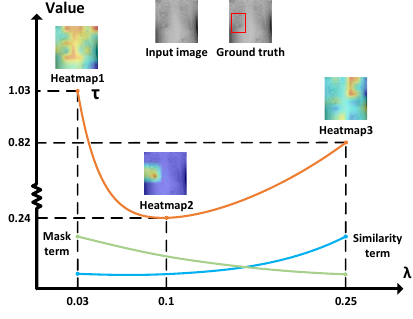}}
	\caption{When the consistency activation loss function reaches the minimum value, the similarity terms, mask terms, $\tau$ values corresponding to different weights $\lambda$ and the visual interpretation heatmap generated by DM at this time. In these heatmaps, the change from blue to red shows a gradual increase in relative importance to network decisions.}
	\label{fig_balance}
\end{figure}

The weight $\lambda$ in the previous DM optimization consistent activation loss (\ref{Ax}) and (\ref{Axb}) was manually set, which resulted in the lack of adaptability of $\lambda$ to different images. In Figure \ref{fig_balance}, the blue line and the green line respectively correspond to the values of the similarity term and mask term after the consistent activation loss is minimized when using different $\lambda$. The heatmaps are the decision saliency maps generated by DM under different $\lambda$. 

If $\lambda$ is small, the value of the similarity term will be optimized to be small after optimizing consistent activation, but the value of the mask term will be optimized to be large. As a result, the trained saliency map retains most of the mask area and cannot remove information redundant areas well, thus failing to accurately locate the lesion, as shown in Heatmap1 in Figure \ref{fig_balance}. If $\lambda$ is large, it will cause the similarity term in the consistent activation loss to be trained larger, which makes the decision-making areas contained in the input image and the mask image inconsistent, resulting in the saliency map being unable to retain the decision-making area and eliminating redundant areas. The effect is as shown in Heatmap3 in Figure \ref{fig_balance}. Therefore, we propose adaptive weight learning to find the most appropriate $\lambda$ to weigh the similarity term and mask term to generate the most effective visual explanation map, such as Heatmap2 in Figure \ref{fig_balance}. Define $\lambda$ as follows:
\begin{equation} \label{lamb_t}
\begin{aligned}
\lambda = \lambda_{0} \lambda_{\nu}
\end{aligned}
\end{equation}

For the training mask vector $\delta_{i}$, $\lambda_{0}$ is initialized to $\frac{Sim(\delta^{0}_{i})}{Mac(\delta^{0}_{i})}$, and $\lambda_{\nu}$ is used as the variable to be learned. The optimal $\lambda$ is obtained by analyzing the optimal $\lambda_{\nu}$ below.

Assume that the saliency map finally generated by DM is $S \in R^{H \times W}$, and we express it in the form of a matrix as follows:
\begin{equation}
\begin{aligned}
S = [s_{i, j}]_{H \times W}
\end{aligned}
\end{equation}
where every element $s_{i, j} \in [0, 1]$, $i \in \{1, 2, ..., H\}$, and $j \in \{1, 2, ..., W\}$.

In saliency maps used to explain decision-making regions in medical images, we think that high-quality saliency maps should have smaller total activation values and a smaller number of extreme values.

Therefore, below we define the total activation value $\sum_{i=1}^{H}\sum_{j=1}^{W}|s_{i,j}|$ and the total discrete extreme value rate $r_{t}$, and use them to measure the quality of the saliency map.

The following introduces the definition of the total discrete extreme rate $r_{t}$. For the horizontal analysis of the saliency map $S$, it can be written as $\{\{s_{i,j}\}^{W}_{j=1}\}^{H}_{i=1}$, that is, it consists of a total of $H$ horizontal discrete point sets $\{s_{i,j}\}^{W}_{j=1}$. We separately calculate the number of horizontal discrete extreme points on these $H$ discrete point sets $\{s_{i,j}\}^{W}_{j=1}$. Obviously, if the saliency map $S$ is continuous, it can be calculated by partial derivatives. But since $S$ is discrete, we define the number of horizontal discrete extreme points $r_{h}$ as the number of triple $(i, u, v)$ that satisfies the inequality group (\ref{r_h}):
\begin{equation} \label{r_h}
\left\{
\begin{aligned}
(s_{i,u} - s_{i, u-1})(s_{i,v} - s_{i, v-1}) < 0, 1 \leq u < v \leq W \\
(s_{i,j+1} - s_{i,j})(s_{i,j+2} - s_{i,j+1}) \geq 0, u \leq j \leq v -3
\end{aligned}
\right.
\end{equation}

In the same way, vertically analyze the saliency map $S$. Divide $S$ into $W$ discrete point sets $\{s_{i,j}\}^{H}_{i=1}$, which can be written as $\{\{s_{i,j}\}^{H}_{i=1}\}^{W}_{j=1}$. We define the number of vertical discrete extreme points $r_{v}$ to be the number of triple $(j, u, v)$ that satisfies the inequality group (\ref{r_v}):
\begin{equation} \label{r_v}
\left\{
\begin{aligned}
(s_{u,j} - s_{u-1, j})(s_{v, j} - s_{v-1, j}) < 0, 1 \leq u < v \leq H \\
(s_{i+1,j} - s_{i,j})(s_{i+2,j} - s_{i+1,j}) \geq 0, u \leq i \leq v -3
\end{aligned}
\right.
\end{equation}

The total discrete extreme rate is defined as follows:
\begin{equation}
\begin{aligned}
r_{t} = \frac{r_{h} + r_{v}}{H(W-2) + W(H-2)}
\end{aligned}
\end{equation}

We want to generate a high-quality saliency map that has a small total discrete extreme rate $r_{t}$ and a small total activation value $\sum_{i=1}^{H}\sum_{j=1}^{W}|s_{i,j}|$. Therefore, we combine the above two to define the evaluation index $\tau$ as follows:
\begin{equation}
\begin{aligned}
\tau = r_{t} \sum_{i=1}^{H}\sum_{j=1}^{W}|s_{i,j}|
\end{aligned}
\end{equation}

\begin{algorithm}[tb]
	\caption{Adaptive Weight Learning}
	\label{alg:algorithm1}
	\textbf{Input}: Weight Candidate Set $\{\lambda_{i}\}^{D}_{i=1}$, Quality Function $Q (\cdot)$, Merge Parameter $\gamma$, Threshold $\theta$, Candidate Number $D$.\\
	\textbf{Output}: Weight $\lambda_{i_{1}}$.
	
	\begin{algorithmic}[1] %[1] enables line numbers
		\STATE $i_{1} \gets \mathop{\arg\min}\limits_{1 \leq i \leq D} Q (\lambda_{i})$ 
		\STATE $i_{2} \gets \mathop{\arg\min}\limits_{i \in \{i_{1}-1, i_{1}+1\}} Q (\lambda_{i})$
		\STATE $\lambda_{i_{3}} \gets \frac{\lambda_{i_{1}} + \gamma \lambda_{i_{2}}}{1 + \gamma}$
		\STATE \textbf{while} $\frac{Q(\lambda_{i_{1}})}{Q(\lambda_{i_{2}})} < \theta$ \textbf{do}
		\STATE \;\;\,\,\,\, \textbf{if} $Q(\lambda_{i_{3}}) < Q(\lambda_{i_{1}})$ \textbf{then}
		\STATE \;\;\,\,\,\,\;\;\,\,\,\,\, $\lambda_{i_{2}} \gets \lambda_{i_{1}}$
		\STATE \;\;\,\,\,\,\;\;\,\,\,\,\, $\lambda_{i_{1}} \gets \lambda_{i_{3}}$
		\STATE \;\;\,\,\,\,\;\;\,\,\,\,\, $\lambda_{i_{3}} \gets \frac{\lambda_{i_{1}} + \gamma \lambda_{i_{2}}}{1 + \gamma}$
		\STATE \;\;\,\,\,\,\;\;\,\,\,\,\, \textbf{elif} $ Q(\lambda_{i_{1}}) \leq Q(\lambda_{i_{3}}) < Q(\lambda_{i_{2}})$ \textbf{then}
		\STATE \;\;\,\,\,\,\;\;\,\,\,\,\,\;\;\,\,\,\,\, $\lambda_{i_{2}} \gets \lambda_{i_{3}}$
		\STATE \;\;\,\,\,\,\;\;\,\,\,\,\,\;\;\,\,\,\,\, $\lambda_{i_{3}} \gets \frac{\lambda_{i_{1}} + \gamma \lambda_{i_{2}}}{1 + \gamma}$
		\STATE \;\;\,\,\,\,\;\;\,\,\,\,\, \textbf{else}
		\STATE \;\;\,\,\,\,\;\;\,\,\,\,\,\;\;\,\,\,\,\, $\lambda_{i_{2}} \gets \lambda_{i_{1}}$
		\STATE \;\;\,\,\,\, \textbf{end if}
		\STATE \textbf{end while}
		\STATE \textbf{return} $\lambda_{i_{1}}$
	\end{algorithmic}
\end{algorithm}

The quality of the saliency map $S$ is evaluated by $\tau$. Obviously, the smaller $\tau$ is, the higher the quality of the saliency map $S$ is. 

According to the above definition, we have the quality function $Q: \lambda_{\nu} \rightarrow \tau$. In Equation (\ref{lamb_t}), since $\lambda_{0}$ is fixed, each $\lambda_{\nu}$ corresponds to a unique $\lambda$ , and each $\lambda$ corresponds to a unique $\tau$. Therefore, we enumerate several $\lambda_{\nu}$ and calculate the corresponding $\tau$ according to the quality function $Q$. Then obtain the optimal $\lambda_{\nu}$ and $\lambda$ through the smallest $\tau$. The pseudocode for calculating $\lambda_{\nu}$ is as shown in Algorithm \ref{alg:algorithm1}.

As shown in Figure \ref{fig_balance}, the minimum $\tau$ is calculated through the above adaptive weight learning to calculate the most appropriate $\lambda$, thereby generating the most effective visual explanation map.

\section{Experiment}

\subsection{Datasets and Baselines}

\subsubsection{Datasets} \

(1) \textbf{iChallenge-PM}\cite{fu2019palm}: iChallenge-PM is a competition jointly organized by Baidu Brain and Zhongshan Ophthalmology Center of Sun Yat-sen University to provide medical data sets for pathological myopia. The dataset covers 400 images in training, validation and test sets. In the test set, some images come with foreground labels about pathological areas.

(2) \textbf{RSNA}\cite{gabruseva2020deep}: This dataset consists of 26,684 forward-view lung CT images and corresponding category labels. This dataset contains 25,684 training data and 1,000 testing data. Some test images provide bounding boxes for pneumonia regions.

(3) \textbf{Covid-19}\cite{chowdhury2020can}: The COVID-19 CXR dataset is specially designed for classification tasks and contains 1,200 COVID-19 positive images and 1,341 normal images.

(4) \textbf{ChinaCXRSet}\cite{jaeger2013automatic,candemir2013lung}: This set of data is preserved by the National Library of Medicine in Maryland, USA. The Shenzhen collection contains 662 CXR images.

(5) \textbf{MontgomerySet}\cite{jaeger2013automatic,candemir2013lung}: The Montgomery group included 138 CXR images (58 tuberculosis, 80 normal) collected under the Montgomery National Tuberculosis Control Program.

\subsubsection{Baselines} \

FeaInfNet was tested and compared with the following multiple baseline models for disease diagnosis capabilities. These models included both interpretable prototype-based networks (ProtoPNet($1 \times 1$), ProtoPNet($2 \times 2$), ProtoPNet($3 \times 3$), TesNet, NP-ProtoPNet, Gen-ProtoPNet, and XProtoNet) as well as uninterpretable backbone networks (VGG19 \cite{simonyan2014very}, ResNet50 \cite{he2016deep}, and DenseNet121 \cite{huang2017densely}).

We compared interpretability using state-of-the-art interpretable networks (ProtoPNet($1 \times 1$), ProtoPNet($2 \times 2$), ProtoPNet($3 \times 3$), NP-ProtoPNet, Gen-ProtoPNet, XProtoNet, and TesNet). We use the above network to generate saliency maps for visualization, and evaluate the ability of different methods to localize lesion regions by comparing these saliency maps with ground-truth labels. The visual interpretability of each model is evaluated through the above comparison.

\subsection{Evaluation}

\subsubsection{Disease Diagnosis Ability}
Diagnose input images into diseased and normal images. Top-1 Accuracy \cite{yan2022discriminative} is used as an evaluation index for the model's disease diagnosis ability. It is calculated as:
\begin{equation}
\begin{aligned}
Accuracy = \frac{N_{correct}}{N_{total}}
\end{aligned}
\end{equation}
where $N_{total}$ is the total number of correctly predicted test images, and $N_{correct}$ is the total number of test images. 

\subsubsection{Lesion Localization Ability}

Interpretable models can generate saliency maps after diagnosing medical images to provide visual explanations of lesion decision-making regions. We calculate the generated saliency map and lesion area ground truth on the following four evaluation indicators: Dice Coefficient \cite{laradji2021weakly}, Positive Predictive Value (PPV) \cite{laradji2021weakly}, Sensitivity \cite{laradji2021weakly}, and Proportion \cite{wang2020score} to evaluate the model's ability to localize the disease.

We perform segmentation according to a set threshold, and set the pixel value of the position in the saliency map whose activation value is lower than the threshold to 0, while the pixel value of the remaining positions is set to 1. Use this method to generate a binary mask map, and then compare the binary mask map with the ground truth to calculate the Dice Coefficient, PPV, and Sensitivity. The formula is:
\begin{equation}
\begin{aligned}
Dice \  Coefficient = \frac{2TP}{FP+2TP+FN} 
\end{aligned}
\end{equation}
\begin{equation}
\begin{aligned}
PPV = \frac{TP}{TP+FP}
\end{aligned}
\end{equation}
\begin{equation}
\begin{aligned}
Sensitivity = \frac{TP}{TP+FN}
\end{aligned}
\end{equation}
where TP, TN, FP, and FN represent true positive, true negative, false positive, and false negative, respectively \cite{singh2021these}.

In order to further measure the saliency map positioning ability, we use Proportion to calculate how much energy in the saliency map falls inside the bounding box area. Its formula is as follows:
\begin{equation}
\begin{aligned}
Proportion = \frac{\sum L^{pos}_{(i,j) \in bbox}}{\sum L^{pos}_{(i,j) \in bbox} + \sum L^{pos}_{(i,j) \notin bbox}}
\end{aligned}
\end{equation}
where the bbox area represents the ground truth foreground or bounding box internal area. $(i, j)$ are the horizontal and vertical coordinates of the saliency map, and $L^{pos}_{(i, j)}$ represents the energy of the corresponding saliency map falling on $(i, j)$ when the model predicts that the input image is a disease image.

\subsection{Experimental Details}

\begin{itemize}
	\item \textbf{Data Augmentation:} Images were rotated, perspectived, sheared, and distorted to generate augmented images. All images were cropped to a size of $224 \times 224$.
	\item \textbf{Hyperparameters:} Various hyperparameters were used, including $H=W=224$, $C=3$, $\xi=0.5$, $\kappa = 2$, $\eta_{1}=\eta_{2}=\eta_{3}=\eta_{4}= 1e-3$, $\epsilon=1e-12$, $T=49$, $C_{1}=128$, and $H_{1}=W_{1}=7$. $\omega$ was the 20th percentile value of saliency map activations sorted in descending order.
	\item \textbf{Number of Prototypes:} Positive shared prototypes were set to a total of 10, while negative prototypes were set to 4 in each region.
	\item \textbf{Training Batches:} The training batch was 20.
	\item \textbf{Hyperparameter Selection:} Cross-validation was used for selection.
	\item \textbf{Optimizer:} The Adam optimizer was used with different learning rates for various layers in the FeaInfNet. The CNN in FeaInfNet consists of the encoding layer of ResNet50, VGG19 or DenseNet121 and shaping layer. The learning rates of the encoding layer and shaping layer are $1e-4$ and $3e-3$ respectively. The learning rates of prototype layer and fully connected layer in the FeaInfNet were set to $1e-4$.
	\item \textbf{Shaping Layer:} A shaping layer consisting of two $1 \times 1$ convolutional layers with ReLU activation was employed.
	\item \textbf{Pre-trained Parameters:} The parameters of the backbone NNs (i.e. the encoding layer in CNN) were initialized with values pre-trained on ImageNet \cite{deng2009imagenet}.
	\item \textbf{Adaptive-DM Parameters:} For dynamic masks learning, $u_{i} = v_{i} = i+5$, and $N_{d}=9$. Each mask vector was trained for $400$ iterations with a learning rate of $2e-3$. For adaptive weight learning, $\theta=1$, $\gamma=1$, $D=45$, and the weight candidate set was $\{ij|i \in \{1e-2,1e-1,1,1e1,1e2\},j \in \{1,2,...,9\}\}$.
	\item \textbf{Prototype Learning:} The first prototype learning was set to the 10th epoch, and subsequent prototype learning was performed every 10 epochs.
	\item \textbf{Upsampling:} Bilinear interpolation was used as the method for upsampling.
	\item \textbf{Binary Mask Threshold:} It was the activation value of the 20th percentile of saliency map activations arranged in descending order.
	\item \textbf{Hardware:} The models were trained on two 2080Ti GPUs using PyTorch.
	\item \textbf{Interpretability Test Samples:} In the RSNA and iChallenge-PM datasets, 200 positive pathology images were randomly selected from each test set as interpretability experimental samples. 
	\item \textbf{Saliency Maps in Interpretability Tests:} A saliency map was represented by the feature vector corresponding to the disease prototype with the highest similarity score predicted by the model. The interpretability of the model was measured by comparing the localization ability of this saliency map.
	\item \textbf{Experimental Subjects:} In order to simplify the diagnosis of multiple diseases, it was equivalent to diagnosing each disease one by one. We learned from multiple independent models, each focusing on a single disease. Therefore, this work only conducted interpretable binary classification experiments on a single disease to evaluate the accuracy and interpretability of the model.
\end{itemize}

\subsection{Explainable Reasoning Process}

\begin{figure*}[!t]
	\centering
	{\includegraphics[width=1.0\linewidth]{{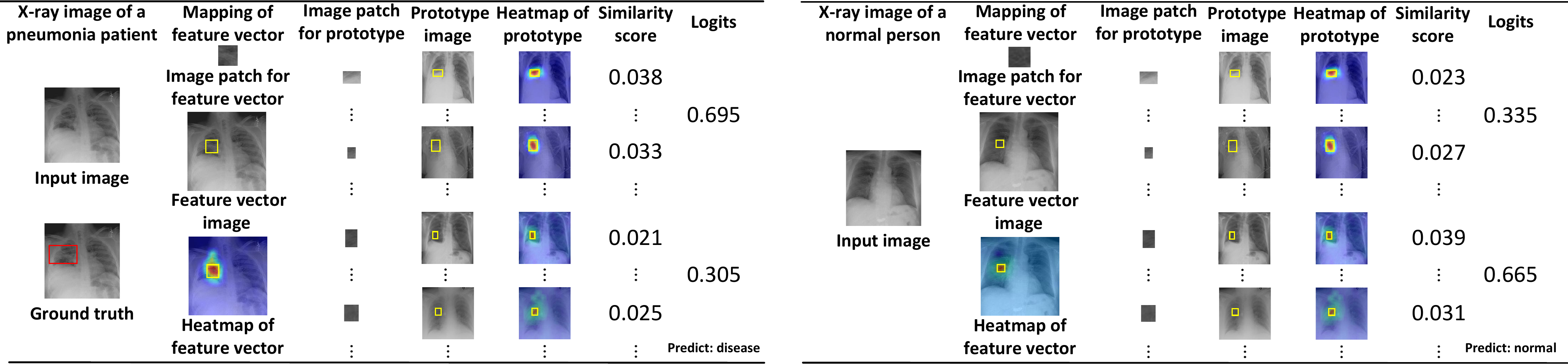}}%
	}
	\caption{The FeaInfNet’s reasoning process in deciding the class of X-ray images of pneumonia patient and normal person.}
	\label{reasong_process}
\end{figure*}

In order to fully demonstrate the interpretability of FeaInfNet with reasoning process in medical image diagnosis. The left and right sides of Figure \ref{reasong_process} visually display the reasoning process of FeaInfNet in diagnosing pneumonia patient images and normal person images respectively. For the pneumonia image on the left, we provide its ground truth. Below we introduce the composition of Figure \ref{reasong_process}:

\begin{itemize}
	\item The ``Mapping of feature vector'' represents the image patch represented by the feature vector in the input image.
	\item The ``Feature vector image'' means a complete image containing feature vectors.
	\item The ``Heatmap of feature vector'' and the ``Heatmap of prototype'' respectively represent the results of Adaptive-DM's visual interpretation of feature vectors and prototypes.
	\item The ``Image patch for prototype'' represents the prototypes that FeaInfNet learns from pneumonia images and normal images in the prototype learning stage, and these prototypes represent patches of corresponding category images in the training set.
	\item The ``Prototype image'' means the complete image corresponding to the prototype in the previous column of the same row.
	\item The ``Similarity score'' indicates how similar the feature vector is to the prototype. The numbers in the upper and lower rows are the similarity scores between the feature vector and the disease prototype and the normal prototype respectively. It consists of 10 positive prototypes and 4 negative prototypes, but due to limited space, only 2 have been written here.
	\item The ``Logits'' are the scores predicted by the network to be pneumonia and normal based on the features, where the upper and lower rows are the scores predicted to be pneumonia and normal respectively.
\end{itemize}

On the left side of Figure \ref{reasong_process}, FeaInfNet encodes the input image into a feature vector (second column). Compare the similarity between the feature vector and the disease prototype and normal prototype (third column) corresponding to the region to obtain the larger disease similarity scores (i.e. 0.038, 0.033 etc.) and the smaller normal similarity scores (i.e. 0.021, 0.025 etc.). 
\begin{equation}
\begin{aligned}
|w^{pos}|= [0.809, 0.700, 2.107, 0.551, 0.846, \\ 
0.776, 0.202, 0.279, 0.097, 0.267]
\end{aligned}
\end{equation}
\begin{equation}
\begin{aligned}
|w^{neg}|= [0.957, 1.708, 1.675, 0.825]
\end{aligned}
\end{equation}
These similarity scores are multiplied by $|w^{pos}|$ and $|w^{neg}|$ respectively, subtracted, and normalized to yield disease and normal logits of 0.695 and 0.305. The input image is finally predicted to be a pneumonia image, and the lesion decision-making area in the input image and the prototype image used as a comparison template are both visualized by Adaptive-DM. The process of predicting a normal image on the right side of Figure \ref{reasong_process} is the same as the prediction process on the left side. Therefore, FeaInfNet provides both interpretability of the reasoning process and visual explanation for disease diagnosis.

\subsection{Network Recognition Ability}

\begin{table*}
	\centering
	\caption{Different types of models, including uninterpretable neural networks and interpretable neural networks with different networks as backbones, were compared for five medical classification datasets to evaluate their performance in terms of recognition accuracy.}
	\resizebox{2.0\columnwidth}{!}{
	\begin{tabular}{@{}lccccccccc@{}}
		\hline
		Datasets & Backbone & ProtoPNet($1 \times 1$) \cite{chen2019looks} & ProtoPNet($2 \times 2$) \cite{chen2019looks} & ProtoPNet($3 \times 3$) \cite{chen2019looks} & TesNet \cite{wang2021interpretable} & NP-ProtoPNet \cite{singh2021these} & Gen-ProtoPNet \cite{singh2021interpretable} & XProtoNet \cite{kim2021xprotonet} & FeaInfNet(ours)\\
		\hline
		Backbone: VGG19\\  		
		RSNA & 79.5 & 77.3 & 80.3 & 81.4  &  80.9   & 77.5 & 78.3 & 80.4 & \textbf{81.9} \\
		iChallenge-PM & 99.0  & 98.0 & 98.5  & 97.3 & 97.3  & 97.3 & 97.5 & 98.3 & \textbf{99.0} \\
		Covid-19 & 98.9  & 97.3 & 97.8  & 97.8  & 96.3 & 97.5 & 97.8 & 97.0 & \textbf{99.4} \\
		ChinaCXRSet & 92.2 & 90.6 & 85.9 & 89.1 & 92.2 & 85.9 & 89.1 & 93.8 & \textbf{95.3} \\
		MontgomerySet & 84.8 & 81.8 & 63.6 & 60.6 & 75.8 & 69.7 & 72.7 & 84.8 & \textbf{87.9} \\
		\hline
		\hline
		Backbone: ResNet50\\  
		RSNA & 78.7 & 74.2 & 80.6 & 81.6 & 78.2 & 79.1 & 79.2 & 77.3 & \textbf{81.7} \\
		iChallenge-PM & 98.5 & 97.5 & 97.5  &  97.3  &  97.8   & 97.3 & 98.3 & 97.5 & \textbf{98.5} \\
		Covid-19 & 97.5  & 95.6 & 97.0 &  97.8  & 97.3 & 96.3 & 95.8 & 94.1 & \textbf{98.0} \\
		ChinaCXRSet & 89.1 & 93.8   &  89.1  & 82.8 & 82.8 & 93.8 & 90.6 & 92.2 & \textbf{95.3} \\
		MontgomerySet & 90.9 & 72.7 & 60.6 &  66.7 &  87.9  & 69.7 & 78.8 & 84.8 & \textbf{90.9} \\
		\hline
		\hline
		Backbone: DenseNet121\\  
		RSNA & 81.5 & 79.4 & 80.7 & 77.1 & 82.3 & 80.7 & 80.1 & 79.5 & \textbf{82.8} \\
		iChallenge-PM & 98.8  & 97.3 & 98.5 &  95.8 & 97.0 & 97.3 & 97.8 & 98.5 & \textbf{98.8} \\
		Covid-19 & 99.0  & 96.0 & 97.5 & 98.3 & 96.8 & 94.6 & 97.1 & 97.8 & \textbf{98.8} \\
		ChinaCXRSet & 93.8 & 92.2 & 84.4 & 84.4 & 87.5 & 90.6 & 93.8 & 93.7 & \textbf{95.5} \\
		MontgomerySet & 87.9 & 75.8 & 63.6 & 63.6 & 72.3 & 81.8 & 78.8 & 87.9 & \textbf{90.9} \\
		\hline
	\end{tabular}}
	\label{images_accuracy}
\end{table*}

As shown in Table \ref{images_accuracy}, we test on five public medical image classification datasets. Compare the classification accuracy of FeaInfNet with uninterpretable backbone networks (second column) and interpretable prototype-based networks (third to ninth columns).

Table \ref{images_accuracy} consists of three major rows, each of which is composed of five small rows, which respectively represent the test results on the corresponding data set. The first, second, and third rows respectively represent the test results when using VGG19, ResNet50, and DenseNet121 as the backbone network. Note that the first to third rows of the backbone in the second column of Table \ref{images_accuracy} represent the uninterpretable networks VGG19, ResNet50, and DenseNet121 respectively. ProtoPNet($r \times r$) in the third to fifth columns follows the ProtoPNet method, and its prototype space size is $r \times r$.

As shown in Table \ref{images_accuracy}, FeaInfNet outperforms other interpretable and uninterpretable baseline models in classification accuracy on various medical datasets. This shows that the feature-based reasoning structure and feature extraction method we proposed are effective in improving the diagnostic accuracy of medical images. FeaInfNet achieves the best classification accuracy while also providing interpretability.

\subsection{Localization Ability of Lesion Area}

\begin{table*}
	\centering
	\caption{Evaluation of the image lesion localization ability of the interpretable neural network using ResNet50 as the backbone network on the RSNA and iChallenge-PM datasets.}
	\begin{tabular}{@{}lcccc|cccc@{}}
		\hline
		Datasets & \multicolumn{4}{c}{RSNA} & \multicolumn{4}{c}{iChallenge-PM} \\
		\hline
		Method & Dice Coefficient & PPV & Sensitivity & Proportion & Dice Coefficient & PPV & Sensitivity & Proportion \\
		\hline
		ProtoPNet($1 \times 1$) \cite{chen2019looks} & 0.116 & 0.181 & 0.101 & 0.166 & 0.138 & 0.078 & 0.197 & 0.171 \\
		ProtoPNet($2 \times 2$) \cite{chen2019looks} & 0.079 & 0.216 & 0.061 & 0.149 & 0.002 & 0.009 & 0.001 & 0.112 \\
		ProtoPNet($3 \times 3$) \cite{chen2019looks} & 0.025 & 0.059 & 0.018 & 0.095 & 0.059 & 0.189 & 0.038 & 0.166 \\
		TesNet \cite{wang2021interpretable} & 0.175 & 0.378 & 0.141 & 0.156 & 0.077 & 0.126 & 0.055 & 0.156 \\
		NP-ProtoPNet \cite{singh2021these} & 0.115 & 0.154 & 0.127 & 0.317 & 0.025 & 0.015 & 0.033 & 0.013 \\
		Gen-ProtoPNet \cite{singh2021interpretable} & 0.175 & 0.240 & 0.175 & 0.224 & 0.113 & 0.061 & 0.168 & 0.155 \\
		XProtoNet \cite{kim2021xprotonet} & 0.107 & 0.148 & 0.108 & 0.186 & 0.051 & 0.028 & 0.088 & 0.092 \\
		FeaInfNet(ours)         & \textbf{0.293} & \textbf{0.379} & \textbf{0.314} & \textbf{0.451} & \textbf{0.314} & \textbf{0.199} & \textbf{0.393} & \textbf{0.549} \\
		\hline
	\end{tabular}
	\label{localization}
\end{table*}

\begin{figure*}[!t]
	\centering
	{\includegraphics[width=1.0\linewidth]{{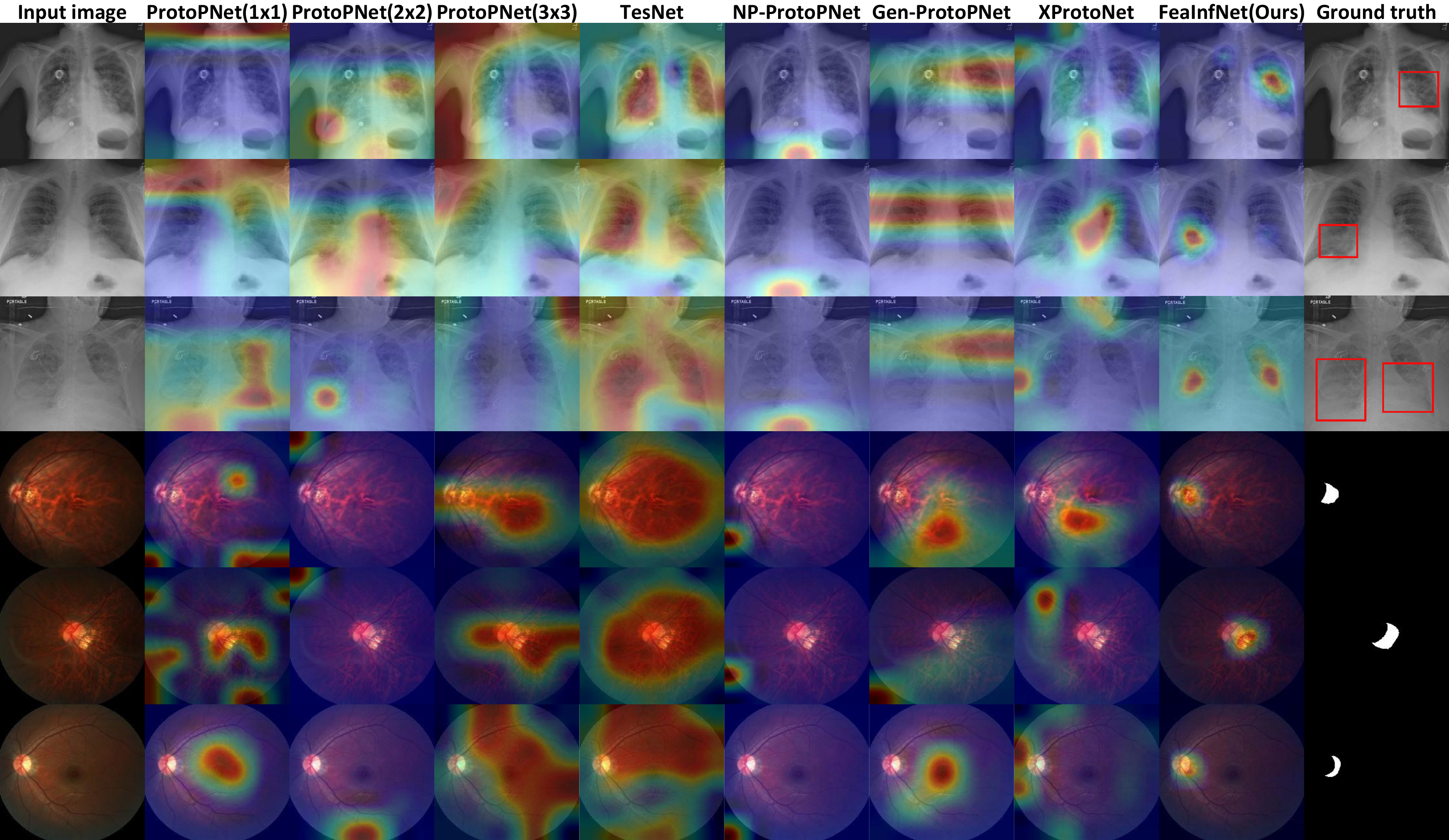}}%
	}
	\caption{The saliency maps generated using various interpretable models for lung and fundus retinal images. The saliency maps generated by each method are normalized to the range [0,1] and visualized using the JET colormap.}
	\label{fig_cam}
\end{figure*}

Both prototype-based neural networks and our proposed FeaInfNet make decisions by comparing feature vectors with prototype similarities. After the model makes a prediction, the feature vectors and prototypes are visualized by generating a saliency map to show which image patches the model focused on when making its decision. Therefore, in this section, we compare the saliency map generated by the previous prototype-based network using upsampled similarity activation to interpret feature vectors with the saliency map generated by FeaInfNet using Adaptive-DM to interpret feature vectors in locating the lesion area. capabilities to evaluate their interpretable performance. In this experiment, each model generates a saliency map of the feature vector corresponding to the prototype with the highest similarity score for comparison.

\textbf{Quantitative evaluation}: Table \ref{localization} compares the RSNA and iChallenge-PM data sets based on four evaluation metrics (Dice Coefficient, PPV, Sensitivity, and Proportion). Compared with state-of-the-art models, our method achieves state-of-the-art performance on all four metrics mentioned above.

\textbf{Qualitative evaluation}: Figure \ref{fig_cam} shows the heatmap generated by mixing the saliency map generated by various methods and the original image. These heatmaps go from blue to red to indicate an increasing probability of focusing on that area when making classification decisions. In the pneumonia images in the first three rows of the last column, the area within the red box is the pneumonia area, and the rest is the normal area. In the last three rows of images, the white area represents the fundus lesion area, and the black area represents the normal area. As can be seen from Figure \ref{fig_cam}, the saliency map generated by the previous prototype-based neural network using the upsampling similarity activation method is not accurate enough for locating the disease area. The FeaInfNet we proposed uses the saliency map generated by Adaptive-DM to accurately locate the disease area for decision-making.

Combining the results of the above quantitative and qualitative evaluations, our method more effectively locates the real pathological regions by interpreting the saliency maps generated by the feature vectors. Therefore, our proposed method has stronger visual interpretability.

\subsection{Learning Process of Adaptive Dynamic Masks}

\begin{figure*}[!t]
	\centering
	{\includegraphics[width=1.0\linewidth]{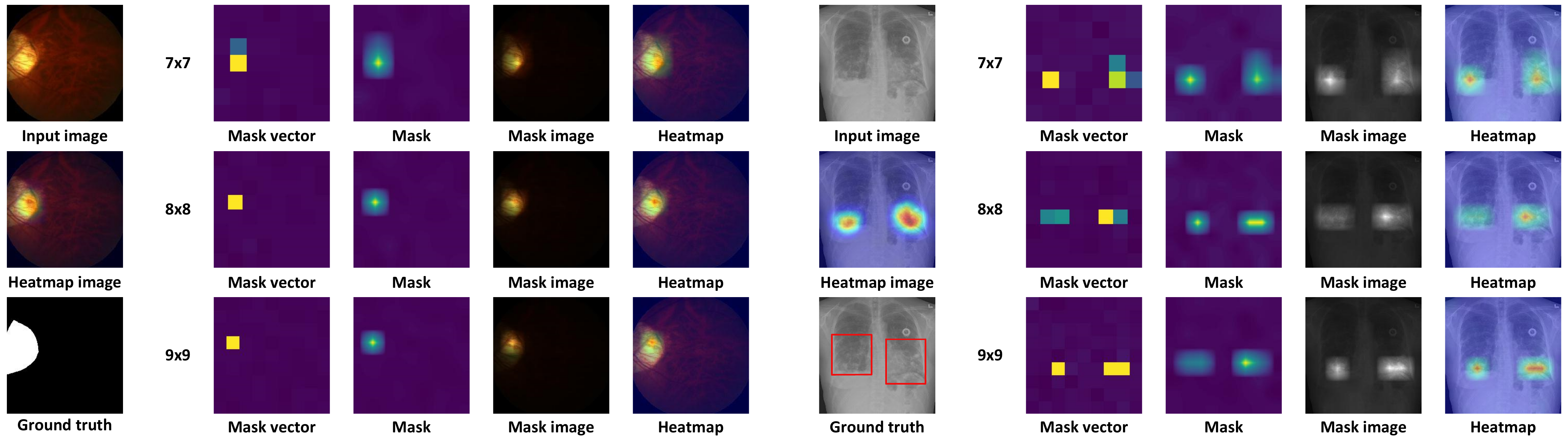}%
	}
	\caption{Two examples of FeaInfNet’s process of learning saliency maps using Adaptive-DM, where a heatmap is obtained by blending the saliency map and the input image using a JET colormap.}
	\label{mask_d}
\end{figure*}

\begin{table}[t]
	\centering
	\caption{Effectiveness of feature-based reasoning. Compare the accuracy of ProtoPNet using feature-based and prototype-based reasoning structures in classifying images in the RSNA dataset under different backbone networks.}
	\resizebox{1.0\columnwidth}{!}{
	\begin{tabular}{@{}lcc@{}}
		\hline
		Method & prototype-based & feature-based \\
		\hline
		ProtoPNet(VGG19) \cite{chen2019looks} & 74.2 & \textbf{78.5}  \\
		ProtoPNet(ResNet50) \cite{chen2019looks} & 77.3 & \textbf{79.9}  \\
		ProtoPNet(DenseNet121) \cite{chen2019looks} & 79.4 & \textbf{79.6}  \\
		\hline
	\end{tabular}}
	\label{ablation_reasoning}
\end{table}

\begin{table}[t]
	\centering
	\caption{Effectiveness of LFM in improving the accuracy of interpretable neural networks on the RSNA dataset. The baseline models for the ablation study are ProtoPNet and FeaInfNet, and the following convolutional neural network is used as the backbone.}
	\resizebox{1.0\columnwidth}{!}{
	\begin{tabular}{@{}lccc@{}}
		\hline
		Method & VGG19 & ResNet50 & DenseNet121 \\
		\hline
		ProtoPNet \cite{chen2019looks} & 74.2 & 77.3 & 79.4 \\
		ProtoPNet+LFM  & \textbf{79.8} & \textbf{80.1} & \textbf{81.7} \\
		\hline
		FeaInfNet & 78.5 & 79.9 & 79.6 \\
		FeaInfNet+LFM  & \textbf{80.6} & \textbf{81.9} & \textbf{82.8} \\
		\hline
	\end{tabular}}
	\label{ablation_lfm}
\end{table}

\begin{table}[t]
	\centering
	\caption{For images in the iChallenge-PM, the evaluation results on Proportion of the saliency maps generated by various interpretable neural networks using different visual interpretation methods.}
	\resizebox{1\columnwidth}{!}{
	\begin{tabular}{lccc}
		\hline
		Method & \multicolumn{1}{c}{Upsampling} & \multicolumn{1}{c}{DM} & \multicolumn{1}{c}{Adaptive-DM(ours)}\\
		\hline
		ProtoPNet \cite{chen2019looks} & 0.171 & 0.223 & \textbf{0.258} \\
		TesNet \cite{wang2021interpretable} & 0.156 & 0.191 & \textbf{0.216} \\
		Gen-ProtoPNet \cite{singh2021interpretable} & 0.155 & 0.182 & \textbf{0.203} \\
		XProtoNet \cite{kim2021xprotonet} & 0.092 & 0.353 & \textbf{0.396} \\
		FeaInfNet(ours) & / & 0.492 & \textbf{0.549} \\
		\hline
	\end{tabular}}
	\label{ablation_adm}
\end{table}

In this section, we visualize the process of Adaptive-DM analyzing FeaInfNet's feature vectors to generate saliency maps to demonstrate the efficacy of this approach.

As shown in Figure \ref{mask_d}, we set up Mask vectors of three sizes ($7 \times 7$, $8 \times 8$, and $9 \times 9$), and generate Mask by upsampling the Mask vector to the original image size. Mask image is the result of element-wise multiplication of Mask and input image, and Heatmap represents the mixed result of Mask and input image. Heatmap image represents the mixture of the input image and the saliency map generated by stacking all upsampled mask vectors, which is also the final result generated by Adaptive-DM.

The $r \times r$-sized mask vector divides the input image into $r \times r$ sub-regions. The smaller $r$ is, the fewer sub-regions are divided, and the larger each region is, the more accurately and widely the mask vector can learn the importance of each sub-region for decision-making. The larger $r$ is, the more sub-regions are divided and the smaller each region is. The mask vector can learn the importance of each sub-region to decision-making in more detail.

As shown in Figure \ref{mask_d}, the saliency map learned by the mask vector of $7 \times 7$ size is relatively rough and can cover a more complete lesion area. The saliency map learned by the $9 \times 9$ size mask vector is relatively fine and can more accurately point out the decision-making area that the network is most concerned about. The saliency map obtained by stacking these upsampled mask vectors can completely and finely pinpoint the pathological regions used by the network for decision-making. The heatmap image shown in Figure \ref{mask_d} combines the completeness of small-size mask vectors and the refinement of large-size mask vectors to generate a complete and accurate saliency map.

The above experiments demonstrate the workflow of Adaptive-DM analyzing FeaInfNet and further verify its feasibility and effectiveness.

\begin{figure}[!t]
	\centerline{\includegraphics[width=\columnwidth]{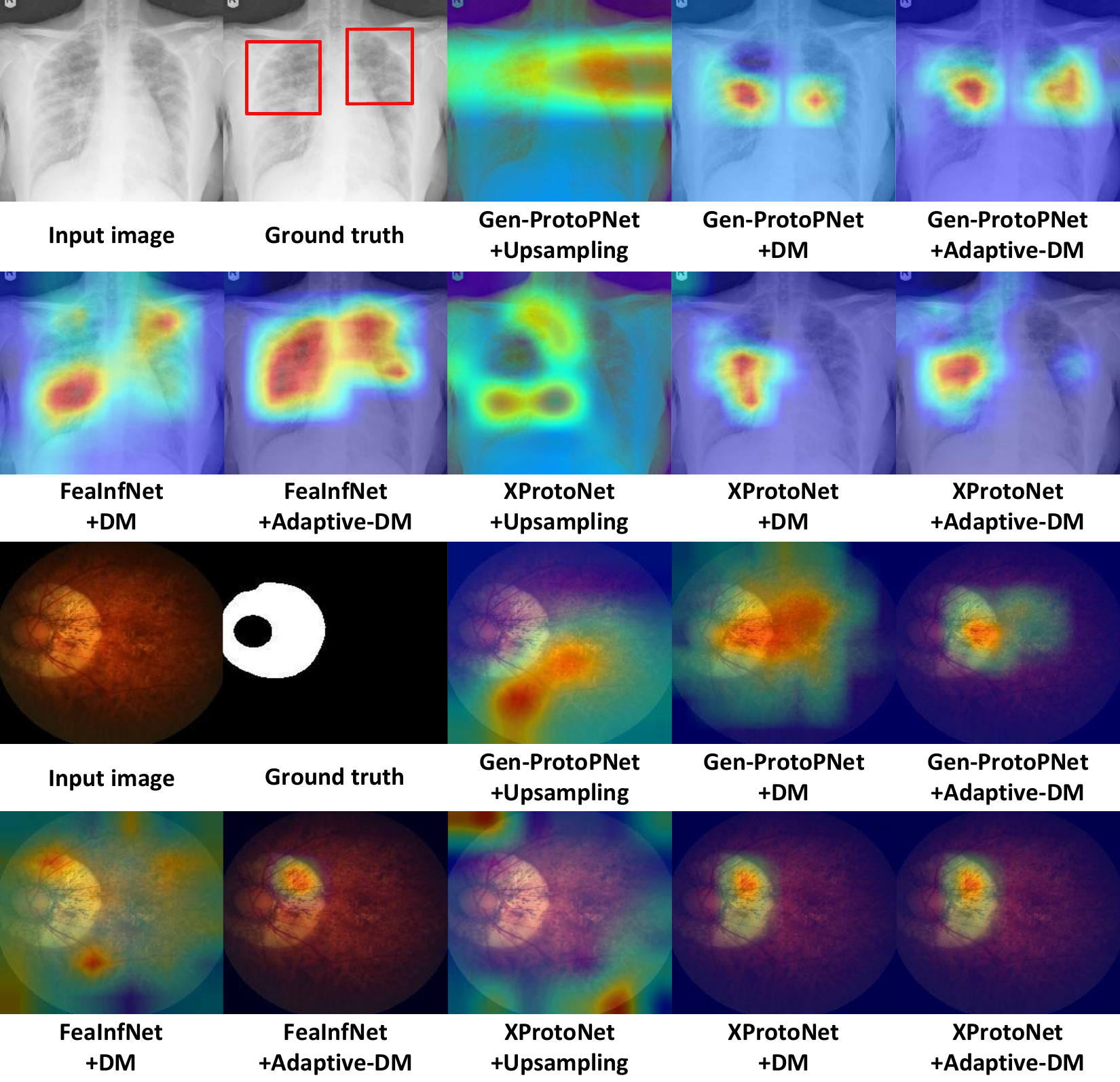}}
	\caption{Visualization results of saliency maps generated by various interpretable neural networks using different visual interpretation methods for images of pneumonia and fundus retinopathy.}
	\label{ablation_cam}
\end{figure}

\begin{figure}[!t]
	\centerline{\includegraphics[width=\columnwidth]{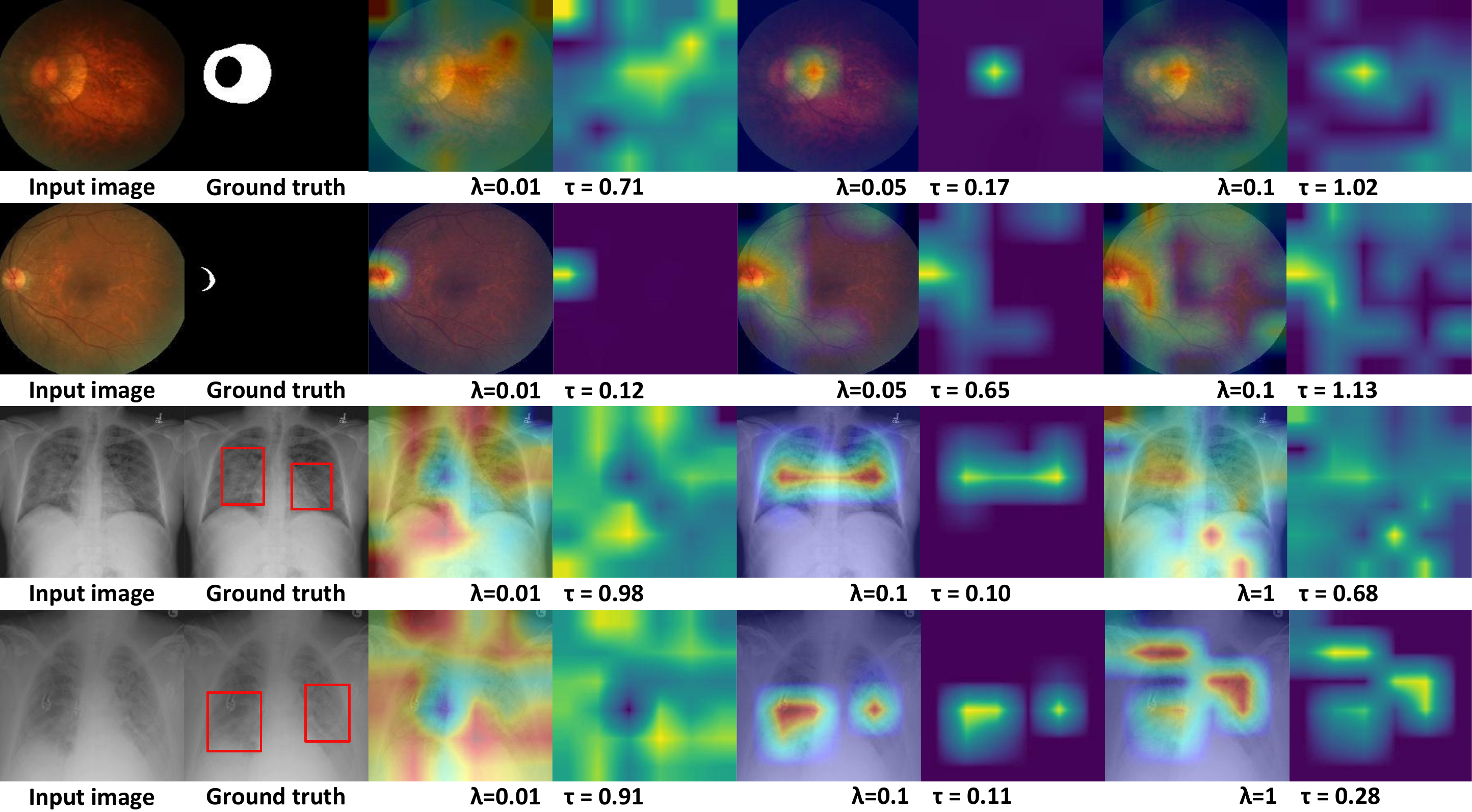}}
	\caption{Under different $\lambda$ values, FeaInfNet uses DM to generate saliency maps and heatmaps of various disease images, and calculates the corresponding $\tau$ values through the quality function.}
	\label{adaptive_lamd}
\end{figure}

\subsection{Ablation Study}

\subsubsection{Effectiveness of Feature-Based Reasoning}

In order to verify whether the feature-based reasoning structure is better than the prototype-based reasoning structure in medical image classification, we chose to conduct an ablation experiment using ProtoPNet as the baseline.

Table \ref{ablation_reasoning} presents the classification accuracy of ProtoPNet on the RSNA data set, using VGG19, ResNet50, and DenseNet121 as the backbone network, respectively using the prototype-based inference structure and the feature-based inference structure. Experimental results show that the accuracy of ProtoPNet using the feature-based reasoning structure is improved by 4.3\%, 2.6\%, and 0.2\% respectively compared with the prototype-based reasoning structure.

Compared with the network with prototype-based reasoning structure, the network with feature-based reasoning structure can more effectively cope with the complex coexistence of categories in medical images by relying only on the characteristics of a single sub-region to make decisions, thus improving the classification accuracy of the network. Experimental results clearly show that networks with feature-based reasoning structure exhibit higher accuracy in medical image classification than networks with prototype-based reasoning structure.

Therefore, our proposed feature-based reasoning structure is effective in medical image diagnosis scenarios.

\subsubsection{Effectiveness of Local Feature Masks}

In order to verify the superiority of local feature masks (LFM) in optimizing network accuracy compared to previous rigid feature vector extraction methods, we conducted ablation studies on ProtoPNet and FeaInfNet, using VGG19, ResNet50, and DenseNet121 as backbone networks.

As shown in Table \ref{ablation_lfm}, using LFM to extract feature vectors significantly improves the accuracy of the above two interpretable networks. Specifically, when ProtoPNet uses VGG19, ResNet50 and DenseNet121 as the backbone, using LFM to extract feature vectors significantly improves ProtoPNet's recognition accuracy on the RSNA dataset by 5.6\%, 2.8\%, and 2.3\%. In addition, FeaInfNet, which uses VGG19, ResNet50 and DenseNet121 as its backbone, showed accuracy improvements of 2.1\%, 2.0\%, and 3.2\% on the RSNA dataset.

Compared with traditional rigid feature extraction methods, LFM significantly enhances the expression efficiency of feature vectors and improves the classification performance of the network by incorporating global information.

\subsubsection{Effectiveness of Adaptive Dynamic Masks}

For the pneumonia images in the RSNA dataset and the fundus lesion images in the iChallenge-PM dataset, we compared the saliency map effects generated by each method to explain the feature vector. Specifically, we visualize the saliency maps generated by Gen-ProtoPNet and XProtoNet using traditional upsampling similarity activation, DM, and Adaptive-DM. We also compared the saliency maps generated by FeaInfNet using DM and Adaptive-DM.

As shown in Table \ref{ablation_adm}, the Proportion comparison of saliency maps generated by multiple interpretable models using upsampling similarity activation, DM, and Adaptive-DM respectively. The results show that in terms of lesion localization performance, the saliency map generated by Adaptive-DM is significantly better than the saliency map generated by DM and upsampling similarity activation method.

In the pneumonia images in the first and second rows of Figure \ref{ablation_cam}, the saliency map generated by Gen-ProtoPNet using traditional upsampling similarity activation covers the pathological area, but the coverage area is very wide, resulting in very inaccurate interpretation. When DM is introduced as an explanation, the saliency map can locate the pathological region relatively well. However, when Adaptive-DM is used, the interpretation area presents a more comprehensive and precise positioning. In XProtoNet, traditional upsampling methods locate errors, while DM and Adaptive-DM can better pinpoint decision-making areas. In FeaInfNet, the decision region explained by Adaptive-DM is more accurate than that explained by DM. In the fundus lesion images in the third and fourth rows, the effect of the saliency map shows that Adaptive-DM is better than DM, and DM is better than the traditional upsampling similarity activations method.

Therefore, Adaptive-DM is able to provide the best visual explanation not only for FeaInfNet, but also for other prototype-based networks. Compared with DM, Adaptive-DM adaptively selects the most appropriate weight $\lambda$ to weigh the similarity term and the mask term by minimizing $\tau$, so that the mask vectors can learn the decision-making area better.

\subsubsection{Effectiveness of Adaptive Weight Learning}

To further demonstrate the effectiveness of our proposed adaptive weight learning. As shown in Figure \ref{adaptive_lamd}, when different $\lambda$ is used as the weight in the consistent activation loss of DM, the trained saliency maps are different. And when DM analyzes different images, the value of $\lambda$ that produces the best results may be different.

In the first two rows of Figure \ref{adaptive_lamd}, we used DM to analyze two fundus images. We show the saliency maps learned by DM for different $\lambda$ values, and the corresponding $\tau$ values of these saliency maps. In the first row, when $\lambda=0.01$, the consistent activation loss pays more attention to the similarity term, causing the mask vector to retain more activation areas, and large areas of the generated saliency map are activated, making the interpretation inaccurate. When $\lambda=0.1$, the consistent activation loss pays too little attention to the similarity items, resulting in the image covered by the mask vector not being well consistent with the original input image result, resulting in the saliency map activation area not being a decision-making area. When $\lambda=0.05$, the importance of the similarity term and the mask term is best weighed, and the generated saliency map well locates the pathological area used for decision-making, and the corresponding $\tau$ is also the smallest at this time. When $\lambda=0.01$ in the second row, $\tau$ can obtain the minimum value of 0.12. The saliency map generated at this time is more accurate in locating the lesion area than the saliency map generated by DM in $\lambda=0.05$ and $\lambda=0.1$.

In the analysis of pneumonia images in the third and fourth lines, by calculating the smallest $\tau$, the corresponding $\lambda$ at this time generates the best saliency map quality.

Therefore, the adaptive weight learning method we proposed calculates the $\tau$ generated under each $\lambda$ to select the smallest $\tau$ to find the most effective $\lambda$, thereby generating a saliency map with the best positioning ability. This also proves the effectiveness of our adaptive weight learning.

\section{Conclusion}
In this research work, we propose FeaInfNet, an interpretable neural network for feature-based reasoning dedicated to medical diagnosis. This innovative model solves the misleading problem that previous prototype-based interpretable models may produce in medical image diagnosis through a feature-based reasoning structure. In addition, we introduce LFM to extract feature vectors, which helps the network learn discriminative features in medical images more effectively by supplementing local feature vectors with some global information. Finally, our proposed Adaptive-DM achieves accurate shaping of the saliency map by effectively balancing the training of similarity terms and mask terms. Provide accurate visual explanations for FeaInfNet and other prototype-based explainable neural networks. Therefore, FeaInfNet is designed to provide a comprehensive solution for medical image diagnosis with both high classification accuracy and strong interpretability. Extensive experiments on five public medical datasets show that our model achieves higher classification accuracy while having the best interpretability compared to other interpretable neural networks. We expect that this research will provide strong support for the future application of deep learning models in the field of medical diagnosis.

\bibliographystyle{IEEEtran}

\bibliography{IEEEabrv,DProtoNet}

\begin{thebibliography}{10}
\providecommand{\url}[1]{#1}
\csname url@samestyle\endcsname
\providecommand{\newblock}{\relax}
\providecommand{\bibinfo}[2]{#2}
\providecommand{\BIBentrySTDinterwordspacing}{\spaceskip=0pt\relax}
\providecommand{\BIBentryALTinterwordstretchfactor}{4}
\providecommand{\BIBentryALTinterwordspacing}{\spaceskip=\fontdimen2\font plus
\BIBentryALTinterwordstretchfactor\fontdimen3\font minus
  \fontdimen4\font\relax}
\providecommand{\BIBforeignlanguage}[2]{{%
\expandafter\ifx\csname l@#1\endcsname\relax
\typeout{** WARNING: IEEEtran.bst: No hyphenation pattern has been}%
\typeout{** loaded for the language `#1'. Using the pattern for}%
\typeout{** the default language instead.}%
\else
\language=\csname l@#1\endcsname
\fi
#2}}
\providecommand{\BIBdecl}{\relax}
\BIBdecl

\bibitem{chandrasekaran2023retinopathy}
R.~Chandrasekaran and B.~Loganathan, ``Retinopathy grading with deep learning
  and wavelet hyper-analytic activations,'' \emph{The Visual Computer},
  vol.~39, no.~7, pp. 2741--2756, 2023.

\bibitem{song2023odspc}
S.~Song, T.~Huang, Q.~Zhu, and H.~Hu, ``Odspc: deep learning-based 3d object
  detection using semantic point cloud,'' \emph{The Visual Computer}, pp.
  1--15, 2023.

\bibitem{bayoudh2021survey}
K.~Bayoudh, R.~Knani, F.~Hamdaoui, and A.~Mtibaa, ``A survey on deep multimodal
  learning for computer vision: advances, trends, applications, and datasets,''
  \emph{The Visual Computer}, pp. 1--32, 2021.

\bibitem{chen2022boosting}
J.~Chen, Z.~Fu, J.~Huang, X.~Hu, and T.~Peng, ``Boosting vision transformer for
  low-resolution borehole image stitching through algebraic multigrid,''
  \emph{The Visual Computer}, vol.~38, no. 9-10, pp. 3191--3203, 2022.

\bibitem{cheng2022contour}
Z.~Cheng, A.~Qu, and X.~He, ``Contour-aware semantic segmentation network with
  spatial attention mechanism for medical image,'' \emph{The Visual Computer},
  pp. 1--14, 2022.

\bibitem{garcia2023secure}
Z.~Garcia-Nonoal, D.~Mata-Mendoza, M.~Cedillo-Hernandez, and
  M.~Nakano-Miyatake, ``Secure management of retinal imaging based on deep
  learning, zero-watermarking and reversible data hiding,'' \emph{The Visual
  Computer}, pp. 1--16, 2023.

\bibitem{gu2020vinet}
D.~Gu, Y.~Li, F.~Jiang, Z.~Wen, S.~Liu, W.~Shi, G.~Lu, and C.~Zhou, ``Vinet: A
  visually interpretable image diagnosis network,'' \emph{IEEE Transactions on
  Multimedia}, vol.~22, no.~7, pp. 1720--1729, 2020.

\bibitem{xi2020integrated}
P.~Xi, H.~Guan, C.~Shu, L.~Borgeat, and R.~Goubran, ``An integrated approach
  for medical abnormality detection using deep patch convolutional neural
  networks,'' \emph{The Visual Computer}, vol.~36, no.~9, pp. 1869--1882, 2020.

\bibitem{lin2023deep}
S.~Lin, A.~Masood, T.~Li, G.~Huang, and R.~Dai, ``Deep learning-enabled
  automatic screening of sle diseases and lr using oct images,'' \emph{The
  Visual Computer}, vol.~39, no.~8, pp. 3259--3269, 2023.

\bibitem{fong2017interpretable}
R.~C. Fong and A.~Vedaldi, ``Interpretable explanations of black boxes by
  meaningful perturbation,'' in \emph{Proceedings of the IEEE international
  conference on computer vision}, 2017, pp. 3429--3437.

\bibitem{yuan2020interpreting}
H.~Yuan, L.~Cai, X.~Hu, J.~Wang, and S.~Ji, ``Interpreting image classifiers by
  generating discrete masks,'' \emph{IEEE Transactions on Pattern Analysis and
  Machine Intelligence}, 2020.

\bibitem{shrikumar2017learning}
A.~Shrikumar, P.~Greenside, and A.~Kundaje, ``Learning important features
  through propagating activation differences,'' in \emph{International
  conference on machine learning}.\hskip 1em plus 0.5em minus 0.4em\relax PMLR,
  2017, pp. 3145--3153.

\bibitem{sundararajan2017axiomatic}
M.~Sundararajan, A.~Taly, and Q.~Yan, ``Axiomatic attribution for deep
  networks,'' in \emph{International conference on machine learning}.\hskip 1em
  plus 0.5em minus 0.4em\relax PMLR, 2017, pp. 3319--3328.

\bibitem{selvaraju2017grad}
R.~R. Selvaraju, M.~Cogswell, A.~Das, R.~Vedantam, D.~Parikh, and D.~Batra,
  ``Grad-cam: Visual explanations from deep networks via gradient-based
  localization,'' in \emph{Proceedings of the IEEE international conference on
  computer vision}, 2017, pp. 618--626.

\bibitem{jiang2021layercam}
P.-T. Jiang, C.-B. Zhang, Q.~Hou, M.-M. Cheng, and Y.~Wei, ``Layercam:
  Exploring hierarchical class activation maps for localization,'' \emph{IEEE
  Transactions on Image Processing}, vol.~30, pp. 5875--5888, 2021.

\bibitem{rudin2019stop}
C.~Rudin, ``Stop explaining black box machine learning models for high stakes
  decisions and use interpretable models instead,'' \emph{Nature Machine
  Intelligence}, vol.~1, no.~5, pp. 206--215, 2019.

\bibitem{rymarczyk2022interpretable}
D.~Rymarczyk, {\L}.~Struski, M.~G{\'o}rszczak, K.~Lewandowska, J.~Tabor, and
  B.~Zieli{\'n}ski, ``Interpretable image classification with differentiable
  prototypes assignment,'' in \emph{Computer Vision--ECCV 2022: 17th European
  Conference, Tel Aviv, Israel, October 23--27, 2022, Proceedings, Part
  XII}.\hskip 1em plus 0.5em minus 0.4em\relax Springer, 2022, pp. 351--368.

\bibitem{donnelly2022deformable}
J.~Donnelly, A.~J. Barnett, and C.~Chen, ``Deformable protopnet: An
  interpretable image classifier using deformable prototypes,'' in
  \emph{Proceedings of the IEEE/CVF Conference on Computer Vision and Pattern
  Recognition}, 2022, pp. 10\,265--10\,275.

\bibitem{chen2019looks}
C.~Chen, O.~Li, D.~Tao, A.~Barnett, C.~Rudin, and J.~K. Su, ``This looks like
  that: deep learning for interpretable image recognition,'' \emph{Advances in
  neural information processing systems}, vol.~32, 2019.

\bibitem{keswani2022proto2proto}
M.~Keswani, S.~Ramakrishnan, N.~Reddy, and V.~N. Balasubramanian,
  ``Proto2proto: Can you recognize the car, the way i do?'' in
  \emph{Proceedings of the IEEE/CVF Conference on Computer Vision and Pattern
  Recognition}, 2022, pp. 10\,233--10\,243.

\bibitem{singh2021interpretable}
G.~Singh and K.-C. Yow, ``An interpretable deep learning model for covid-19
  detection with chest x-ray images,'' \emph{IEEE Access}, vol.~9, pp.
  85\,198--85\,208, 2021.

\bibitem{nauta2021neural}
M.~Nauta, R.~Van~Bree, and C.~Seifert, ``Neural prototype trees for
  interpretable fine-grained image recognition,'' in \emph{Proceedings of the
  IEEE/CVF Conference on Computer Vision and Pattern Recognition}, 2021, pp.
  14\,933--14\,943.

\bibitem{rymarczyk2021protopshare}
D.~Rymarczyk, {\L}.~Struski, J.~Tabor, and B.~Zieli{\'n}ski, ``Protopshare:
  Prototypical parts sharing for similarity discovery in interpretable image
  classification,'' in \emph{Proceedings of the 27th ACM SIGKDD Conference on
  Knowledge Discovery \& Data Mining}, 2021, pp. 1420--1430.

\bibitem{10314012}
Y.~Peng, L.~He, D.~Hu, Y.~Liu, L.~Yang, and S.~Shang, ``Hierarchical dynamic
  masks for visual explanation of neural networks,'' \emph{IEEE Transactions on
  Multimedia}, pp. 1--15, 2023.

\bibitem{zhou2016learning}
B.~Zhou, A.~Khosla, A.~Lapedriza, A.~Oliva, and A.~Torralba, ``Learning deep
  features for discriminative localization,'' in \emph{Proceedings of the IEEE
  conference on computer vision and pattern recognition}, 2016, pp. 2921--2929.

\bibitem{rajpurkar2018deep}
P.~Rajpurkar, J.~Irvin, R.~L. Ball, K.~Zhu, B.~Yang, H.~Mehta, T.~Duan,
  D.~Ding, A.~Bagul, C.~P. Langlotz \emph{et~al.}, ``Deep learning for chest
  radiograph diagnosis: A retrospective comparison of the chexnext algorithm to
  practicing radiologists,'' \emph{PLoS medicine}, vol.~15, no.~11, p.
  e1002686, 2018.

\bibitem{lin2020covid}
T.-C. Lin and H.-C. Lee, ``Covid-19 chest radiography images analysis based on
  integration of image preprocess, guided grad-cam, machine learning and risk
  management,'' in \emph{Proceedings of the 4th International Conference on
  Medical and Health Informatics}, 2020, pp. 281--288.

\bibitem{lopatina2020investigation}
A.~Lopatina, S.~Ropele, R.~Sibgatulin, J.~R. Reichenbach, and D.~G{\"u}llmar,
  ``Investigation of deep-learning-driven identification of multiple sclerosis
  patients based on susceptibility-weighted images using relevance analysis,''
  \emph{Frontiers in neuroscience}, vol.~14, p. 609468, 2020.

\bibitem{sayres2019using}
R.~Sayres, A.~Taly, E.~Rahimy, K.~Blumer, D.~Coz, N.~Hammel, J.~Krause,
  A.~Narayanaswamy, Z.~Rastegar, D.~Wu \emph{et~al.}, ``Using a deep learning
  algorithm and integrated gradients explanation to assist grading for diabetic
  retinopathy,'' \emph{Ophthalmology}, vol. 126, no.~4, pp. 552--564, 2019.

\bibitem{kim2021xprotonet}
E.~Kim, S.~Kim, M.~Seo, and S.~Yoon, ``Xprotonet: diagnosis in chest
  radiography with global and local explanations,'' in \emph{Proceedings of the
  IEEE/CVF conference on computer vision and pattern recognition}, 2021, pp.
  15\,719--15\,728.

\bibitem{singh2021these}
G.~Singh and K.-C. Yow, ``These do not look like those: An interpretable deep
  learning model for image recognition,'' \emph{IEEE Access}, vol.~9, pp.
  41\,482--41\,493, 2021.

\bibitem{wang2021interpretable}
J.~Wang, H.~Liu, X.~Wang, and L.~Jing, ``Interpretable image recognition by
  constructing transparent embedding space,'' in \emph{Proceedings of the
  IEEE/CVF International Conference on Computer Vision}, 2021, pp. 895--904.

\bibitem{wang2020non}
Z.~Wang, N.~Zou, D.~Shen, and S.~Ji, ``Non-local u-nets for biomedical image
  segmentation,'' in \emph{Proceedings of the AAAI conference on artificial
  intelligence}, vol.~34, no.~04, 2020, pp. 6315--6322.

\bibitem{yang2022focal}
Z.~Yang, Z.~Li, X.~Jiang, Y.~Gong, Z.~Yuan, D.~Zhao, and C.~Yuan, ``Focal and
  global knowledge distillation for detectors,'' in \emph{Proceedings of the
  IEEE/CVF Conference on Computer Vision and Pattern Recognition}, 2022, pp.
  4643--4652.

\bibitem{fu2019palm}
H.~Fu, F.~Li, J.~I. Orlando, H.~Bogunovic, X.~Sun, J.~Liao, Y.~Xu, S.~Zhang,
  and X.~Zhang, ``Palm: Pathologic myopia challenge,'' \emph{IEEE Dataport},
  2019.

\bibitem{gabruseva2020deep}
T.~Gabruseva, D.~Poplavskiy, and A.~Kalinin, ``Deep learning for automatic
  pneumonia detection,'' in \emph{Proceedings of the IEEE/CVF conference on
  computer vision and pattern recognition workshops}, 2020, pp. 350--351.

\bibitem{chowdhury2020can}
M.~E. Chowdhury, T.~Rahman, A.~Khandakar, R.~Mazhar, M.~A. Kadir, Z.~B. Mahbub,
  K.~R. Islam, M.~S. Khan, A.~Iqbal, N.~Al~Emadi \emph{et~al.}, ``Can ai help
  in screening viral and covid-19 pneumonia?'' \emph{IEEE Access}, vol.~8, pp.
  132\,665--132\,676, 2020.

\bibitem{jaeger2013automatic}
S.~Jaeger, A.~Karargyris, S.~Candemir, L.~Folio, J.~Siegelman, F.~Callaghan,
  Z.~Xue, K.~Palaniappan, R.~K. Singh, S.~Antani \emph{et~al.}, ``Automatic
  tuberculosis screening using chest radiographs,'' \emph{IEEE transactions on
  medical imaging}, vol.~33, no.~2, pp. 233--245, 2013.

\bibitem{candemir2013lung}
S.~Candemir, S.~Jaeger, K.~Palaniappan, J.~P. Musco, R.~K. Singh, Z.~Xue,
  A.~Karargyris, S.~Antani, G.~Thoma, and C.~J. McDonald, ``Lung segmentation
  in chest radiographs using anatomical atlases with nonrigid registration,''
  \emph{IEEE transactions on medical imaging}, vol.~33, no.~2, pp. 577--590,
  2013.

\bibitem{simonyan2014very}
K.~Simonyan and A.~Zisserman, ``Very deep convolutional networks for
  large-scale image recognition,'' \emph{arXiv preprint arXiv:1409.1556}, 2014.

\bibitem{he2016deep}
K.~He, X.~Zhang, S.~Ren, and J.~Sun, ``Deep residual learning for image
  recognition,'' in \emph{Proceedings of the IEEE conference on computer vision
  and pattern recognition}, 2016, pp. 770--778.

\bibitem{huang2017densely}
G.~Huang, Z.~Liu, L.~Van Der~Maaten, and K.~Q. Weinberger, ``Densely connected
  convolutional networks,'' in \emph{Proceedings of the IEEE conference on
  computer vision and pattern recognition}, 2017, pp. 4700--4708.

\bibitem{yan2022discriminative}
T.~Yan, H.~Li, B.~Sun, Z.~Wang, and Z.~Luo, ``Discriminative feature mining and
  enhancement network for low-resolution fine-grained image recognition,''
  \emph{IEEE Transactions on Circuits and Systems for Video Technology},
  vol.~32, no.~8, pp. 5319--5330, 2022.

\bibitem{laradji2021weakly}
I.~Laradji, P.~Rodriguez, O.~Manas, K.~Lensink, M.~Law, L.~Kurzman, W.~Parker,
  D.~Vazquez, and D.~Nowrouzezahrai, ``A weakly supervised consistency-based
  learning method for covid-19 segmentation in ct images,'' in
  \emph{Proceedings of the IEEE/CVF Winter Conference on Applications of
  Computer Vision}, 2021, pp. 2453--2462.

\bibitem{wang2020score}
H.~Wang, Z.~Wang, M.~Du, F.~Yang, Z.~Zhang, S.~Ding, P.~Mardziel, and X.~Hu,
  ``Score-cam: Score-weighted visual explanations for convolutional neural
  networks,'' in \emph{Proceedings of the IEEE/CVF conference on computer
  vision and pattern recognition workshops}, 2020, pp. 24--25.

\bibitem{deng2009imagenet}
J.~Deng, W.~Dong, R.~Socher, L.-J. Li, K.~Li, and L.~Fei-Fei, ``Imagenet: A
  large-scale hierarchical image database,'' in \emph{2009 IEEE conference on
  computer vision and pattern recognition}.\hskip 1em plus 0.5em minus
  0.4em\relax IEEE, 2009, pp. 248--255.

\end{thebibliography}

\vfill

\end{document}